\documentclass{article}

\PassOptionsToPackage{authoryear, sort&compress, round}{natbib}

\usepackage[finalws]{neurips_2025}

\usepackage[utf8]{inputenc} 
\usepackage[T1]{fontenc}    
\usepackage{hyperref}       
\usepackage{url}            
\usepackage{booktabs}       
\usepackage{amsfonts}       
\usepackage{nicefrac}       
\usepackage{microtype}      
\usepackage{xcolor}         
\usepackage{graphicx}
\usepackage{subcaption}
\usepackage{placeins}
\usepackage{listings}
\usepackage{multirow}
\usepackage{soul}
\usepackage{array}
\usepackage{wrapfig}

\newcommand{\arslan}[1]{#1}

\title{On the generalization of language models from in-context learning and finetuning: a controlled study}

\author{%
  Andrew K. Lampinen\thanks{Equal contribution.} \\
  Google DeepMind\\
  Moutain View, CA\\
  \texttt{lampinen@google.com}\\
  \And
  Arslan Chaudry$^*$\\
  Google DeepMind\\
  Moutain View, CA\\
  \texttt{arslanch@google.com}\\  
  \And
  Stephanie C. Y. Chan$^*$\\
  Google DeepMind\\
  San Francisco, CA\\
  \texttt{scychan@google.com}\\
  \And
  Cody Wild\\
  Google DeepMind\\
  Moutain View, CA\\
  \And
  Diane Wan\\
  Google DeepMind\\
  Seattle, WA\\
  \And
  Alex Ku\\ 
  Google DeepMind\\
  Mountain View, CA\\
  \And
  J\"{o}rg Bornschein\\
  Google DeepMind\\
  London, UK\\
  \And
  Razvan Pascanu\\ 
  Google DeepMind\\
  London, UK\\
  \And
  Murray Shanahan\\
  Google DeepMind\\
  London, UK\\
  \And
  James L. McClelland\\
  Google DeepMind \& Stanford University\\
  Mountain View, CA\\
}

\begin{document}

\maketitle

\begin{abstract}
Large language models exhibit exciting capabilities, yet can show surprisingly narrow generalization\footnote{By generalization we mean simple logical deductions of the trained knowledge which is the fundamental aspect of reasoning.} from finetuning. E.g. they can fail to generalize to simple reversals of relations they are trained on, or fail to make simple logical deductions based on trained information. These failures to generalize factual information from fine-tuning can significantly hinder the reasoning capabilities of these models. On the other hand, language models’ in-context learning shows different inductive biases and deductive reasoning capabilities. Here, we explore these differences in generalization and deductive reasoning between in-context- and fine-tuning-based learning. To do so, we constructed several novel datasets to evaluate and improve models' abilities to make generalizations over factual information from novel data.
These datasets are designed to create clean tests of generalization, by isolating the knowledge in the dataset from that in pretraining. 
We expose pretrained large models to controlled subsets of the information in these datasets --- either in context, or through fine-tuning --- and evaluate their performance on test sets that require various types of generalization. We find overall that in data-matched settings, in-context learning can generalize several types of inferences more flexibly than fine-tuning (though we also find some qualifications of prior findings, such as cases when fine-tuning can generalize to reversals embedded in a larger structure of knowledge). 
We build on these findings to propose a method to enable improved generalization from fine-tuning: adding in-context reasoning traces to finetuning data. We show that this method improves generalization and deductive reasoning across various splits of our datasets and other benchmarks. Our results have implications for understanding the different forms of reasoning afforded by different modes of learning in language models, and practically improving their performance.
\end{abstract}  

\section*{Introduction} \label{sec:introduction}

Language models (LMs) pretrained on huge corpora of internet text become efficient in-context learners; they can generalize from a few examples of a task to answer new instances \citep{brown2020language,team2023gemini}. Pretrained LMs can also be fine-tuned for downstream tasks using relatively few examples---though achieving good generalization from fine-tuning often requires hundreds, or even thousands of examples \citep[e.g.][]{kirstain2022few,vieira2024much}. Indeed, the generalization from fine-tuning on a particular example can be surprisingly limited; for example, LMs fine-tuned on a statement like ``B's mother is A'' fail to generalize to answer ``Who is A's son?'' (\citealp{berglund2024reversal}; cf. \citealp{allen2025physics3_2}). \arslan{These failures of simple logical deduction of finetuned knowledge can hinder reasoning capabilities of LMs}. However, LMs can readily reason through questions about these types of reverse relations in context \citep[e.g.][]{lampinen2024language}. Furthermore, transformers show different inductive biases when generalizing from weights vs. from context \citep[cf.][]{chan2022transformers,shen2023pretrained,russin2024human}. How do patterns of generalization differ between in-context reasoning and fine-tuning? What are the implications for how we should adapt models for new tasks or information? In this paper, we explore these questions.

To do so, we construct controlled synthetic datasets of factual knowledge. 
We design these datasets to have complex and self-consistent structures, while avoiding any overlap with knowledge that may be present in pretraining corpora. We create train and test splits of these datasets that involve different types of generalization: reversals or chaining multiple logical inferences into a syllogism. 
We then evaluate how well large pretrained language models generalize to these test sets through fine-tuning, or through in-context learning---via placing the entire training set (or large subsets thereof) in context. These results show that models can use long contexts effectively to generalize better than gradient-based learning. We also explore various methods of improving generalization, such as data augmentation. Overall, we find that across various datasets, in-context learning (ICL) generalizes better than finetuning. However, finetuning generalization can be improved, and in fact can even exceed ICL generalization, by spending more train time compute to augment the training dataset using in-context reasoning.

Our contributions are as follows:
\begin{itemize}
    \item We study the distinct patterns of factual generalization that pretrained LMs exhibit from in-context learning and finetuning.
    \item We find that in-context learning over the entire training dataset often generalizes better than finetuning, when evaluated on systematic holdouts like reversals, syllogistic inferences, and other logical deductions.
    \item We propose to bridge this gap via dataset augmentation---prompting an LM to reason about the data in-context, and then adding these reasoning traces to the training set.
    \item We show that dataset augmentations can bridge the gap to yield improved generalization from fine-tuning. 
    \item We also propose a fine-tuning method that breaks up correlations among sentences, amplifying the benefits of augmentation.
\end{itemize}
\section*{Datasets} \label{sec:datasets}

We evaluate learning and generalization across several datasets that are crafted to isolate different aspects of generalization, or to situate it within a broader set of learning challenges. We also draw on datasets from prior work.

\subsection*{Reversal curse paper}

We first use the reversal dataset proposed by \citet{berglund2024reversal} that contains one-sentence descriptions of fictional celebrities. Dataset examples can have the celebrity name (e.g. `Daphne Barrington') preceding the description (e.g. `the director of "A Journey Through Time."') or vice-versa. 
The dataset includes two independent sets of celebrities -- `A' and `B'. Following \citet{berglund2024reversal}, we finetune the model on celebrities from the set `A' with names preceding the description, and from set `B' where names and descriptions appear in both orders but in separate examples. Overall, the train set consists of 3600 examples. The test set evaluates whether the model can infer the celebrity names from the set `A' given only their descriptions. To add distractors in test examples, we include three randomly selected incorrect celebrity names in the list of options to score.

\subsection*{Simple reversals and syllogisms}

We then test our methods with two simple datasets\footnote{These datasets are adapted from the nonsense NLI and Syllogism datasets from \citealt{lampinen2024language}.} that contain independent examples of reversal and syllogistic reasoning. 

\textbf{Simple reversals:} Each training example consists of a brief preamble (e.g. ``Did you know that'') followed by a single statement comparing two entities, e.g. ``femp are more dangerous than glon.'' There are a hundred such comparisons provided (with the comparison words sampled from a set of 28, e.g. ``brighter'', ``heavier,''). Each comparison is repeated across 10 different training examples, each time paired with a randomly sampled preamble. The test set consists of forced choices between the correct reversal and a contradictory relation, e.g.: ``glon are {less/more} dangerous than femp.'' Note that the incorrect answer is a tempting alternative, as it uses the same words as the trained statement (just reordered), which means that a model relying on simple features like bag-of-words would tend to prefer the incorrect answer.

\textbf{Simple syllogisms:} There are 69 training examples. Each example consists of a preamble (which specifies the entities under discussion), followed by two statements that form a syllogism. For example: ``The relationships between glon, troff and yomp are: All glon are yomp. All troff are glon.'' The test examples evaluate whether the model makes a correct inference from that syllogistic form. They provide a preamble specifying two entities from the training example, and then score all possible statements about the relationship between the two entities involving the quantifiers and relations in the dataset. Following \citet{lampinen2024language}, we omit the partial negative form (some X are not Y); thus, there are six such possible statements (the product of ``all'', ``some'', and ``no'' together with the two directions each of those relations could take). For example, the preamble for the test example corresponding to the syllogism above would be ``The relationships between yomp and troff are:'' and the correct answer would be ``All troff are yomp.''

\subsection*{Semantic structure benchmark}

\begin{wrapfigure}{R}{0.45\textwidth}
    \includegraphics[width=\linewidth,trim={0 5pt 0 0},clip]{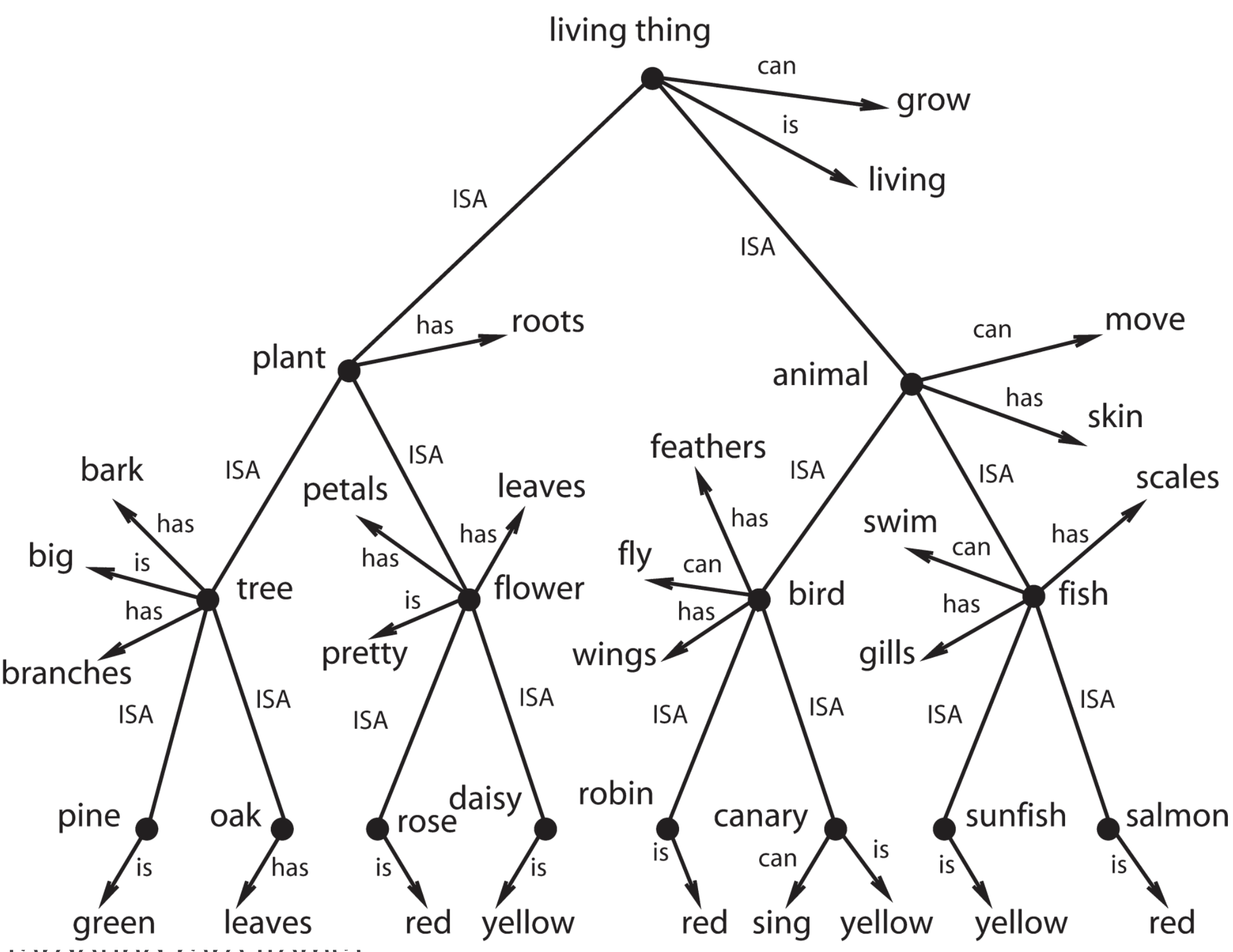}
    \caption{A semantic structure with a hierarchy of properties and relations. (Reproduced with permission from \citet{rogers2008precis}.)}
    \label{fig:semantic_structure}
\end{wrapfigure}

The benchmark is built around a relational semantic hierarchy that allows deductive inferences and abstractions. This hierarchy is based on real-world categories and relations, and draws inspiration from prior work in cognitive science that has studied learning of rich semantic hierarchies (see Fig. \ref{fig:semantic_structure} for an intuitive example that composes part of the real-world structure). We similarly create a hierarchical knowledge structure involving 110 categories of animals and objects, and their properties and relations. Unlike the original structure, it is not balanced in branching factors or depth, and properties are similarly asymmetrically distributed across entities (1-6 per category, plus inherited ones)---making its distributional statistics more naturalistic (see Appx. \ref{appx:methods:datasets} for details). This underlying structure is derived from the real world. 

In order to make the structure novel to the pretrained models, however, we replace all nouns, adjectives, and verbs with nonsense terms. This removes the potential overlap with pretraining data, thus ensuring that the data properties follow certain real-world distributional features, but are not contaminated. 

The use of nonsense terms potentially poses tokenization challenges; however, we ameliorate these by generating short 4-5 letter nonsense words using plausible combinations of phonemes for English \citep[sampled via][]{gao2023scope}. Moreover, we show in the next sections that in practice the models can readily make inferences about these nonsensical entities in context, so use of unfamiliar words is not the primary bottleneck on model performance.

\textbf{Train set:} For training, we assemble train-set facts about this semantic hierarchy into short vaguely-Wikipedia-like synthetic articles with varied formatting and styles, as well as some QA examples (to ensure that fine-tuning on the data doesn’t degrade question answering capabilities). We ensure that all facts that are necessary for the test questions (below) are presented in at least one training document. However, some facts will be redundantly presented across multiple documents. We create a total of 2200 training documents of 4-20 sentences in length.

\textbf{Test sets:} Within the semantic structure, our test sets accommodate reversals of relations (gruds [dogs] are a type of abmes [mammals] => abmes include gruds; cf. \citealp{berglund2024reversal}), syllogism-like deductive inferences (e.g. gruds [dogs] are abmes [mammals]; abmes are rony [warm-blooded] => gruds are rony), and longer deductions. Specifically, we focus on the following splits (in rough difficulty order):

\begin{itemize}
\item \emph{Rephrasings} of trained facts that don’t change the direction of wording: used for validation that the material is learned.
\item \emph{Reversals} of trained facts, i.e. the two entities are in reverse order, and the relational words are adjusted accordingly.
\item \emph{Syllogisms} over trained facts; that is, statements that can be logically inferred by linking two (implicitly) quantified statements in the training corpus, using an entity that overlaps between them.
\item \emph{Category holdouts}: only one fact about a category is present in training: what its parent category is. All possible inferences from that fact are tested. This overlaps with some aspects of the syllogism splits, except that: 1) the information about the target category is more strictly limited, thus limiting other cues that could aid generalization; and 2) inferring the fact may require chaining more than two statements together.
\end{itemize}

When creating the evaluation questions, we choose difficult distractors for the incorrect answers, by choosing entities or properties that have the target relationship to a \emph{different} entity in the dataset. For example, if the correct answer is ``gruds are rony'' one of the distractors might be ``gruds are zept'' where there is some other valid statement in the dataset such as ``telk are zept.'' Thus, it is not possible to guess simply by the local context of the words which answer is correct.

We evaluate the models both with and without few-shot QA examples taken from the validation set in the prompt; we find relatively similar results across these conditions so we collapse across them in our main analyses.

\section*{Methods}


\subsection*{Evaluation}

We perform evaluation using multiple-choice likelihood scoring. We do not provide the answer choices in context. We generally construct the datasets to provide challenging alternative answers (see above).

\textbf{In-context evaluation:}
To perform full-dataset in-context evaluation, we concatenate the documents in a training dataset and provide them as context to the (instruction-tuned) model. We then prompt the model to use that context to answer the question. We randomly subsample the documents by a factor of 8x when evaluating models in-context on the largest datasets, as there is some redundancy in the datasets, and we find that models experience interference as the context length scales.

\subsection*{Fine-tuning}

Our fine-tuning experiments mainly involve tuning Gemini 1.5 Flash \citep{team2024gemini} on our datasets. We generally fine-tune with batch size 8 or 16 and learning rate \(3 \cdot 10^{-4}\) for 200-1000 steps (depending on dataset size and loss).

\subsection*{Dataset augmentation}

Our key approach to dataset augmentation is to take advantage of the generalization of models during in-context reasoning in order to improve the coverage of the fine-tuning dataset. We approach this via several methods, but all chiefly aim at the goal of spending training time compute for in-context inference to create more finetuning data, in order to improve generalization out-of-context (i.e., without requiring additional information in context) at test time.

Specifically, we consider two types of augmentation: a local strategy that tries to increase the flexibility with which particular pieces of information can be used, and global strategies that attempt to relate between different pieces of information. Each of these strategies uses distinct context and prompts (Appx. \ref{appx:methods:prompts}).

\textbf{Local (sentence) augmentation:} We prompt an LM to augment each training data point (e.g. sentence) to enhance the flexibility with which the model encodes it. Our prompt includes examples of rephrasings and reversals.

\textbf{Global (document) augmentation:} We concatenate the full training dataset as context, then provide a particular document and prompt the model to generate inferences by linking between that document and the rest of the documents in context. This results in a longer reasoning trace of relevant inferences.

\subsection*{Sentence-splitting}

Some datasets, such as the fictional celebrities dataset by \citet{berglund2024reversal} and our semantic structure dataset, contain documents comprising multiple sentences that are logically or semantically linked. We find that splitting these documents at sentence-level, into multiple finetuning examples, leads to a much improved finetuning performance -- even after accounting for the total dataset size and gradient steps. We explore two ways of splitting a document: 1) {\em Independent Splitting}: wherein all \(n\) sentences of a document are split independently into \(n\) separate training examples, 2) {\em Cumulative Splitting}: wherein an \(n\)-sentence document is split into \(n\) examples, with the \(i\)th example contains all sentences up to the \(i\)th sentence. We analyze the effect of sentence splitting on the model generalization in appendix~\ref{appx:experiments:sent_split}. In the following sections, we assume independent sentence splitting unless stated otherwise.

\section*{Experiments} \label{sec:experiments}

\textbf{Reversal Curse:} 
\begin{wrapfigure}{R}{0.5\textwidth}
    \vspace{-1em}
    \centering
    \includegraphics[width=\linewidth]{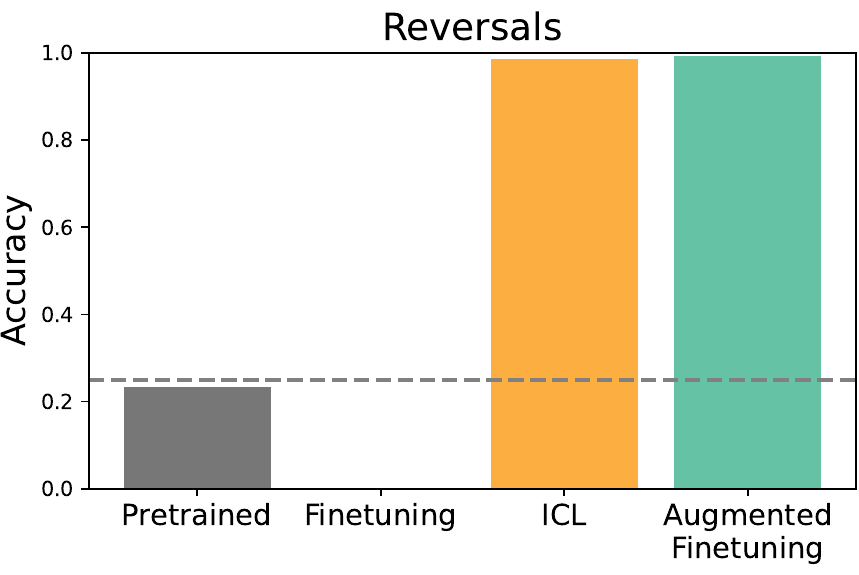}
    \caption{Reversal curse paper results.}
    \label{fig:rcp_dataset}
\end{wrapfigure}

In Fig. \ref{fig:rcp_dataset}, we first explore generalization in the context of the reversal curse phenomenon and dataset released by \citet{berglund2024reversal}. We replicate the finding that finetuning on the forward directions does not produce generalization to the reversals. The authors of that work noted briefly that in-context the models could perform better at this task. We study this more systematically by presenting the entire dataset in context, and find that the model performs nearly at ceiling on the reversals---thus providing a strong demonstration of the benefits of in-context learning over finetuning. Finetuning with data augmented with in-context inferences yields similarly high test performance. Simple finetuning, on the other hand, has near zero accuracy as the finetuned model always prefers those (incorrect) celebrity name completions that were seen as targets during training, regardless of the context; both ICL and Augmented Finetuning significantly outpeform base finetuning (both \(t>149\), \(p<10^{-16}\)). Finally, a pretrained model performs near chance on the test set, indicating a lack of contamination.

\begin{figure*}[bhp]
    \centering
    \begin{subfigure}{0.5\textwidth}
        \centering
        \includegraphics[width=\linewidth]{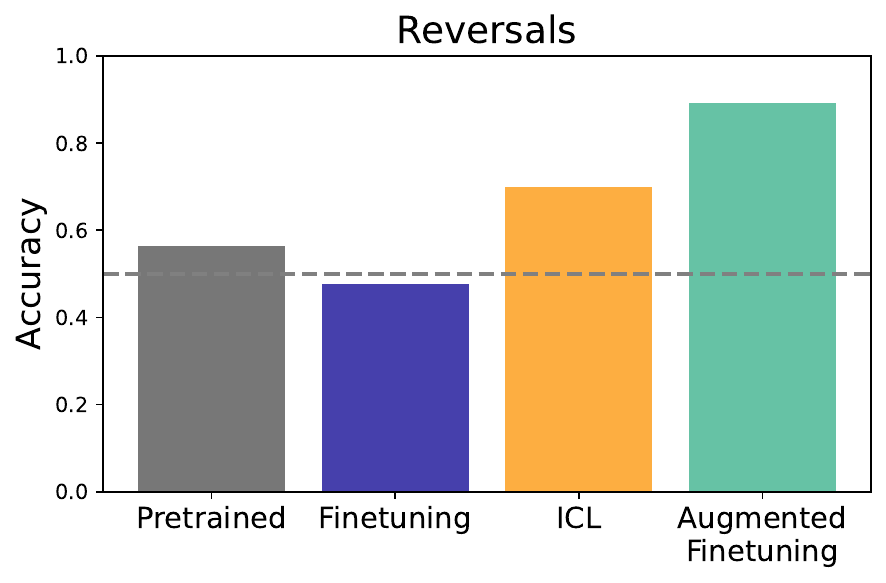}
        \label{fig:simple_datasets:reversal}
    \end{subfigure}%
    \begin{subfigure}{0.5\textwidth}
        \centering
        \includegraphics[width=\linewidth]{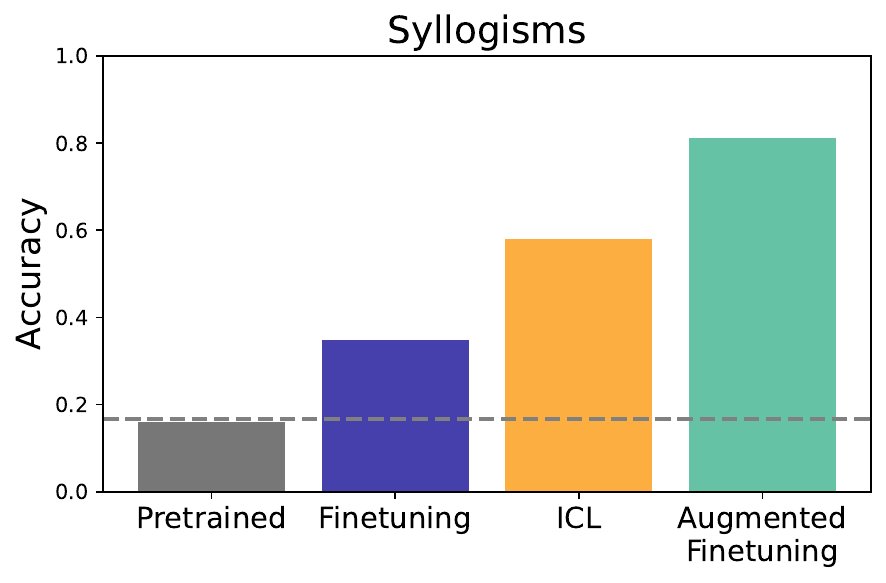}
        \label{fig:simple_datasets:syllogisms}
    \end{subfigure}%
    
    \caption{On our simple reversal (left) and syllogism (right) datasets, in-context learning outperforms finetuning. Moreover, augmenting the fine-tuning dataset produces strong improvements in model performance. Pretrained models perform near chance, showing that the datasets are not guessable based on superficial features.}
    \label{fig:simple_datasets}
\end{figure*}

\textbf{Simple nonsense reversals:}
We then correspondingly test the models on the simple nonsense version of our reversal dataset (Fig~\ref{fig:simple_datasets}, left). We find a weaker, but still noticeable, benefit of ICL over finetuning in this setting, and again stronger benefits of augmented finetuning (respectively \(ts=3.36,7.25\), \(ps < 10^{-3}, 10^{-11}\)). The difference in benefits compared to the experiments above may be due to differences in the plausibility of the relationships in question, e.g. the possibility that nonsense words interfere to some degree with the model's reasoning over longer contexts (see Appx. \ref{appx:experiments:icl_nonsensification}).

\textbf{Simple syllogisms:}
We next test the simple syllogisms dataset (Fig~\ref{fig:simple_datasets}, right). Again, the pretrained model performs at chance, indicating a lack of contamination. Finetuning on the dataset does yield some non-trivial above-chance generalization; perhaps because for certain flavors of syllogism the logical inferences are compatible with simpler linguistic patterns --- e.g., tuning on sequences like ``All X include Y.'' ``All Y include Z.'' might make the model more likely to predict ``All X include Z'' from pure associations. However, following simpler associative patterns is not valid for most syllogistic forms, e.g. if the universal quantifier is replaced by the existential in the examples above. Perhaps because of this, in-context learning yields much stronger performance; using in-context learning to augment the training dataset further improves performance overall (respectively \(ts=2.83,6.30\), \(ps < 0.01,10^{-8}\)).

\subsection*{Semantic structure benchmark}
\begin{figure*}[ht!]
    \centering
    \begin{subfigure}{0.45\textwidth}
        \centering
        \includegraphics[width=\linewidth]{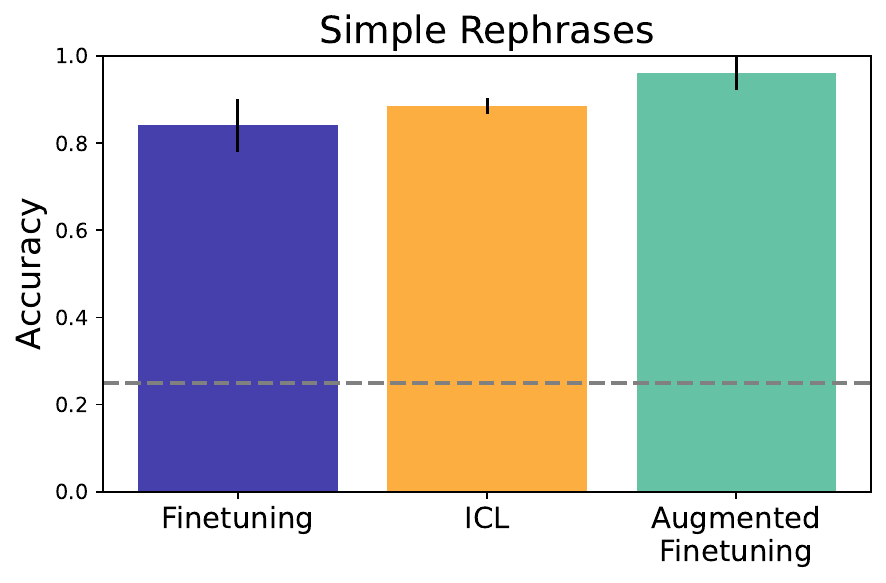}
        \label{fig:semantic:reversal}
    \end{subfigure}%
    \begin{subfigure}{0.45\textwidth}
        \centering
        \includegraphics[width=\linewidth]{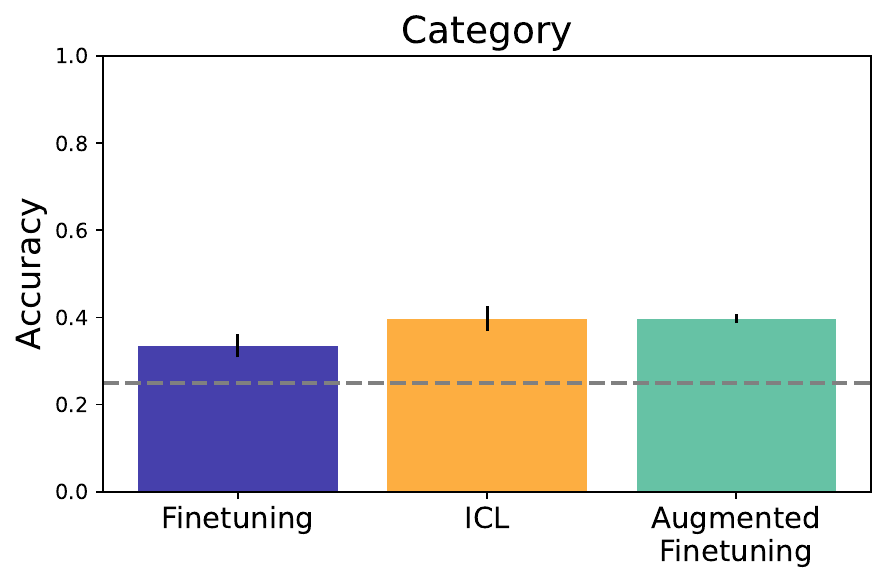}
        \label{fig:semantic:syllogisms}
    \end{subfigure}%
    \vspace{0.1cm}
    \begin{subfigure}{0.45\textwidth}
        \centering
        \includegraphics[width=\linewidth]{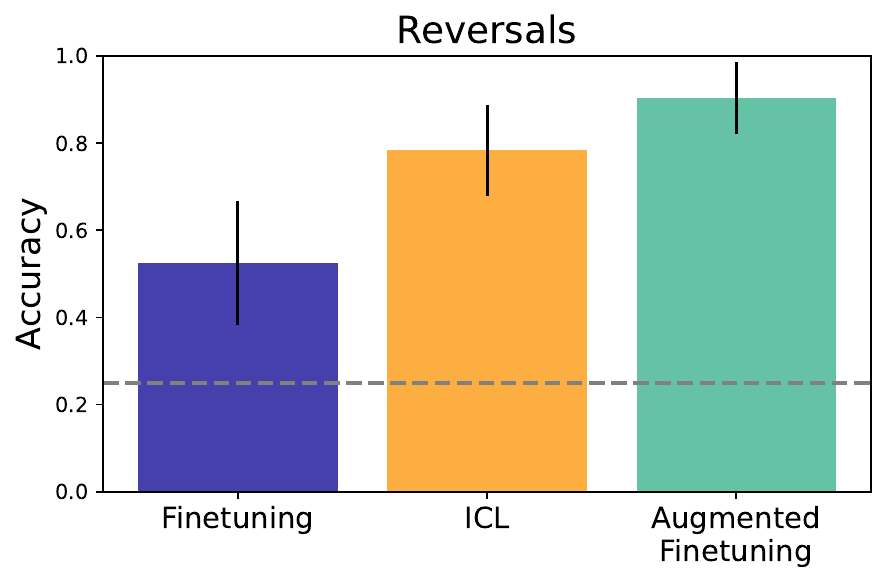}
        \label{fig:semantic:reversal}
    \end{subfigure}%
    \begin{subfigure}{0.45\textwidth}
        \centering
        \includegraphics[width=\linewidth]{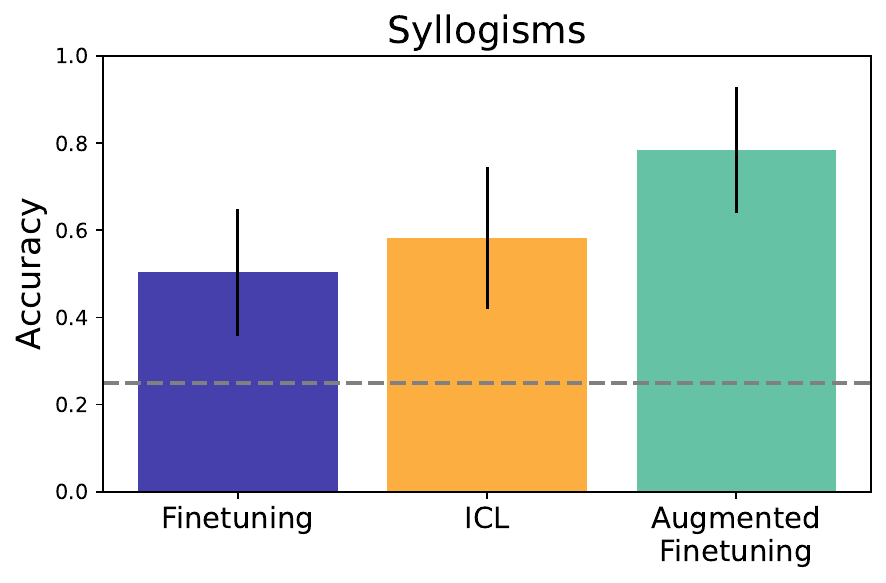}
        \label{fig:semantic:syllogisms}
    \end{subfigure}%
    \caption{On the more richly structured semantic dataset, in-context learning still moderately outperforms finetuning. Furthermore, augmentation continues to show benefit for generalization from finetuning --- even in rephrased questions about trained facts that do not involve reversal (top left). However some generalization splits, such as the category-level holdout, remain very challenging. (Error bars are standard errors computed over task subsets featuring different types of the inferences in question, e.g. reversals of property relations vs. reversals of category inclusion relations.)}
    \label{fig:semantic_results}
\end{figure*}

We then test the semantic structure benchmark (Fig. \ref{fig:semantic_results}), which integrates multiple kinds of generalization in a richer learning setting. In this setting, we evaluate performance on: a) rephrased training statements (that preserve the direction of the relation) denoted as `train' in the figure, b) reversals, c) syllogistic inferences, and d) propositions about held out categories. Across these settings, we find an overall benefit of ICL over finetuning, though the magnitude of this benefit varies depending on the split. We find some improvements in generalization to even rephrased information from training, and more dramatic improvements on reversals and syllogisms. However, the category-holdout split remains difficult, and improvements from ICL are minimal. Furthermore, we continue to find that augmenting the finetuning data with in-context inferences improves performance from finetuning, in many cases outperforming even ICL. Most improvements are statistically significant (\(ts>2.79\), \(p<0.01\)) except for ICL on Syllogisms and both on Category show only marginal improvements (\(1.95>ts>1.93\), \(p=0.05\)). (N.B., in this setting we \emph{do not} use sentence splitting for the finetuning baseline, as we find it impairs performance and we want to compare to the strongest baselines; results with sentence-splitting for the baseline, along with other ablations, can be found in Appx. \ref{appx:experiments:prompts_comparison}.)

Our results in this section also highlight an important nuance to prior results on the reversal curse \citep{berglund2024reversal}. When the information being tested forms part of a broader coherent structure of knowledge (such as our semantic structure), finetuning alone does exhibit \emph{some} above-chance generalization to reversals. This generalization may be due to the fact that other information in the training set can support the reversed conclusion; e.g. if the reversal were from ``birds include eagles'' to ``eagles are a type of bird,'' but the training set also includes statements such as ``eagles have wings,'' that information may enable an alternative route to inferring the reversal (associatively if not logically). Nevertheless, in-context learning and augmented finetuning continue to perform substantially better than finetuning alone in most cases. 

\textbf{Process knowledge:}
We also performed some preliminary experiments exploring learning and generalization of process-type (rather than semantic) knowledge, i.e. the ability to apply novel procedures to inputs. Specifically, we focused on executing a simple pseudo-mathematical procedure that we call ``derivatoids'', which transforms math-like expressions in a manner similar to computing a derivative. We created a train-test compositional generalization split in which certain combinations of rules are seen in training (again, either in context or via finetuning), but other combinations are held out for testing. The overall pattern of results is generally consistent with those above. ICL outperforms finetuning, especially in the low-shot regime, and augmentations improve finetuning performance. However, process knowledge requires distinct evaluation methods, and the effects correspondingly appear to be driven by distinct factors compared to the semantic knowledge experiments above; for brevity, we present the experiments and results in full detail in Appendix \ref{appx:experiments:process}.
\section*{Related work} \label{sec:related_work}

\textbf{In-context learning:} A variety of works have explored in-context learning \citep{brown2020language,dong2022survey,lampinen2024broader}. Several works have studied patterns in the learning and generalization of ICL, either empirically \citep{garg2022can,chan2022data}, mechanistically \citep{olsson2022context}, or theoretically \citep{xie2023explanation,zhang2024trained,elmoznino2024context}. Several recent works have found that even scaling to hundreds or thousands of in-context examples can improve LM performance on challenging datasets \citep{agarwal2024many,anil2024many,bertsch2024context}. Our work similarly relies on the ability of models to learn from many documents in context in order to generate the inferences for augmentation. Our work differs from prior works in focusing more on generalization of broader types of information in context, beyond few-shot tasks \citep[cf.][]{lampinen2024broader}.

\textbf{Out-of-context learning:} Several other papers \citep[e.g.][]{berglund2023taken,meinke2023tell} have considered how models can generalize ``out-of-context''---that is, how they can use information that is not directly included in the prompt in a flexible way at test time. Our results may connect to these, e.g. in the patterns of partial generalization seen even with finetuning on the semantic structure dataset. However, we generally do not find reliable use of out-of-context information as in-context---i.e., in-context learning tends to outperform finetuning. 

\textbf{Factual learning \& generalization:} A variety of works have studied how language models learn and generalize factual information, both in pretraining \citep[e.g.][]{allen2024physics3_1,allen2025physics3_2,zucchet2025language} and during finetuning \citep[e.g.][]{berglund2024reversal,gekhman2024does}. Many of these works have focused on failures to generalize like the ``Reversal Curse'' \citep{berglund2024reversal}, while others looked into increases in hallucination as LLMs are pretrained \citep{zucchet2025language} or finetuned on new knowledge \citep{gekhman2024does}. However, these works have generally not performed data-matched experiments comparing the generalization of in-context learning and finetuning. 

\textbf{Data augmentation:} A wide variety of works have explored how LLMs can be used to augment data for improved performance from small or narrow datasets, e.g. to improve zero-shot task performance \citep{maini2024rephrasing} or to improve cross-lingual generalization \citep[e.g.][]{whitehouse2023llm}. \citet{ding2024data} reviews some of this broader literature. There have also been targeted attempts to fix specific issues like the reversal curse with hardcoded augmentations \citep{golovneva2024reverse}. A closely related work by \citet{akyurek2024deductive} proposes ``deductive closure training'' to improve coverage via prompting language models to generate deductive inferences from training documents. \citet{padmanabhan2024propagating} similarly propose generating continuations and then distilling them into the model. Several concurrent works \citet{chen2024reverse,ruan2025reasoning} suggest that having LMs generate additional reasoning directions, and using these to augment their training data, can improve performance on reasoning tasks.  \citet{kujanpaa2024knowledge} propose prompt distillation, which distills from larger prompts containing privileged information, and find that this improves generalization from knowledge in finetuning. \citet{yang2024synthetic} use language models to extract entities in the training documents, and generate synthetic data reasoning about links between these entities; like our results, they find this method of ``rearranging'' knowledge in synthetic data helps to improve downstream performance. Finally, a concurrent work \citep{park2025textit} finds similar benefits to ours of in-context learning over finetuning, and likewise proposes augmenting data to overcome it. Our experiments show that in controlled settings, without the possibility of dataset contamination, similar approaches to augmenting small finetuning datasets can yield improved generalization---and relate the performance to that achieved through more basic finetuning and in-context learning.

\textbf{Synthetic data:} An equally broad literature has explored other applications of synthetic data to improving LM performance; see \citet{liu2024best} for a recent survey. Earlier works considered hand-designed synthetic data targeted to improve generalization using domain knowledge of areas like linguistics or mathematics \citep[e.g.][]{pratapa2018language,wu2021lime}. More recent approaches have focused on generating the data directly from language models, either filtering with ground-truth scores \citep{zelikman2022star} or simply through self-consistency \citep{wang2023self,huang2023large}. While a recent prominent article argued that training on synthetic data leads models to collapse \citep{shumailov2024ai}, other works have noted that in settings where synthetic data is mixed with the original data performance continues to improve \citep{gerstgrasser2024model}. We correspondingly find that in our setting, rather than being harmful, synthesizing augmented data with the models consistently improves performance (even on rephrased information from the training split in the semantic structure dataset). These results contribute to the ongoing discussion of how incorporating synthetic data affects model performance.

\section*{Discussion} \label{sec:discussion}

In this paper, we performed controlled experiments on how language models generalize various types of novel information from in-context learning and finetuning. Overall, we find that the models generalize better on average along several dimensions from in-context learning. Using in-context reasoning to augment the finetuning dataset can exploit the complementary benefits of both to yield better performance. 

\textbf{The distinct inductive biases of in-context learning and finetuning:}
A variety of works have studied the inductive biases of in-context learning. One common theme has been to emphasize that in-context learning can approximate gradient descent \citep[e.g.][]{vonoswald2023transformers}, in settings where that behavior is optimal. However, a variety of other works have found that the inductive biases of in-context learning can vary depending on factors such as dataset diversity \citep{raventos2024pretraining} or model scale \citep{wei2023larger}. Indeed, \citet{elmoznino2024context} argue that in-context learning has an ``Occam's razor''-like bias towards minimizing complexity, which enhances its generalization. Several works have explicitly noted the distinct inductive biases of in-context vs. in-weights learning (e.g. ``rule-based'' vs ``exemplar-based'' generalization) \citep{chan2022transformers,shen2023pretrained,awadalla2022exploring}---while others have argued that they are more similar than they might seem \citep{mosbach-etal-2023-shot}. Our work contributes to this line of findings, by moving beyond the typical input-output tasks considered in these works, to consider other types of knowledge---and illustrates some potential utility of these differences, by showing how the more flexible generalization of in-context learning can be exploited to improve the generalization of fine-tuning.

\textbf{Accessible information and learning by thinking:} \citet{lombrozo2024learning} highlights how ``learning by thinking'' is a unifying theme across cognitive science and recent advances in AI---a system can acquire new information and skills purely through computation, without further inputs. Lombrozo highlights that while superficially this may seem paradoxical---information cannot be created---this further computation can increase the \emph{accessibility} of information and thus improve performance in practice. This argument parallels the theoretical one made by \citet{xu2020theory} on how computation can increase the accessibility of information. Our use of in-context reasoning to improve finetuning performance beyond the original data follows this pattern. For example, the information about reversals and syllogisms is always hidden within the data, but finetuning on the in-context inferences make this information more explicit and thus more readily accessible at test time. 

\textbf{Train-time inference scaling:} 
Recently, various works have begun to explore test-time inference scaling to improve performance \citep[e.g.][]{jaech2024openai,guo2025deepseek}.
These findings complement prior studies exploring how scaling training compute (e.g. via larger models or more data) improves performance \citep[e.g.][]{kaplan2020scaling,hoffmann2022training}. Our results illustrate that scaling train-time compute through in-context inference methods can likewise help to improve some aspects of model generalization.

\textbf{Limitations:} Our work suffers from several limitations. First, our main experiments rely on nonsense words and implausible operations. Although these counterfactual tasks allow us to avoid the possibility of dataset contamination, they may interfere with the performance of the models to some extent. For example, preliminary experiments (Appx. \ref{appx:experiments:icl_nonsensification}) suggest that the ICL performance of the models on the Reversal Curse dataset degrades if the names are replaced with nonsense. Thus, tasks with more plausible entities may see greater benefits from ICL. It's possible that finetuning makes the nonsense terms more familiar, and that this contributes to the gap between augmented finetuning and ICL. However, in that case we likely \emph{underestimate} the gap between ICL and base finetuning (as the latter would effectively benefit from increased ``familiarity'' of the nonsense entities). Second, we have not experimented with other language models, which would enhance the generality of our results. However, since the individual phenomena we build upon --- e.g. the reversal curse when finetuning \citep{berglund2024reversal} vs. the ability to do reversals in context \citep[e.g.][]{lampinen2024language} --- have been documented across multiple models, we believe it is reasonable to extrapolate our results to other settings. However, future work should study more carefully the differences in how models learn and generalize in these settings, including newer reasoning models \citep[e.g.][]{guo2025deepseek}.  

\subsection*{Conclusions}
We have explored the generalization of in-context learning and finetuning when LMs are exposed to various types of novel information structures. We find notable differences---often with ICL generalizing better for certain types of inferences---and propose ways of achieving better performance by adding in-context reasoning traces to augment finetuning data. We hope this work will contribute to the science of understanding learning, reasoning, and generalization in foundation models, and the practicalities of adapting them to downstream tasks.
\section*{Acknowledgements}
We would like to thank the following people for helpful discussions: Sridhar Thiagarajan, Mike Mozer, Amal Rannen-Triki, Andrey Zhmoginov, Preethi Lahoti, Dilan Gorur, Johannes von Oswald. We would also like to thank Shakir Mohamed, Dipanjan Das, Raia Hadsell, Slav Petrov, Andrew Dai, Ruibo Liu, Tom Kwiatkowski, and Lukas Dixon for their support.   

\bibliography{main}

\clearpage
\appendix
\onecolumn

\section{Supplemental methods} \label{appx:methods}

\subsection{Prompts} \label{appx:methods:prompts}

In this section we provide the prompts used for the augmentations. The relevant portions of the prompt were formatted with the corresponding content.

\begin{lstlisting}[basicstyle=\footnotesize]
LOCAL_PROMPT = (
    'Please generate possible novel statements and rephrasings that can be '
    'inferred from each sentence on its own. Even a simple sentence has '
    'some alternative phrasings that are logically equivalent. Please '
    'only use logic and language to draw your conclusions, regardless of '
    'whether the entities in question actually exist.\n',
    'Statement: trillips are taller than zax.',
    'Inferences: trillips have greater height than zax. zax are shorter'
    ' than trillips. zax have lower heights than trillips.',
    'Statement: Note: Engineering is simpler than science.',
    'Inferences: Science is more complex than engineering. Engineering is '
    'less complex than science. Engineering is not as complex as science.',
    'Statement: "{text_to_augment}"',
)

GLOBAL_PROMPT = (
    'I am going to give you a bunch of documents, and then please use them as '
    'context to reason through all the logical consequences you can produce '
    'from a final target document. First, here are the source documents:\n\n'
    '{full_context}\n\n'
    'Now, please use that context to help me rephrase this document or reason '
    'through the consequences of the statements it contains. Please state all '
    'consequences as explicitly as possible, following the format of the '
    'source documents, and be as complete as possible.\n'
    '{target_document}\n'
)
\end{lstlisting}

For some of our ablation experiments we also used a document-level prompt without the global context:

\begin{lstlisting}[basicstyle=\footnotesize]
DOCUMENT_PROMPT = (
    'Please generate possible novel propositions that can be logically '
    'inferred from each document on its own. Even a simple sentence has '
    'some alternative phrasings that are logically equivalent, and '
    'combining sentences can often yield new logical inferences. Please '
    'only use logic and language to draw your conclusions, regardless of '
    'whether the entities in question actually exist.\n'
)


\end{lstlisting}

\subsection{Datasets} \label{appx:methods:datasets}

\textbf{Semantic structure}

The full tree structure used to derive the benchmark has 110 entities structured into a tree with depth of up to 7, with branching factors ranging from 1-8, and numbers of node-specific properties ranging from 1-6. 

The (non-nonsense) relations used, e.g. ``are a type of'' were converted into natural language by a range of templates (per relation). In turn, these were embedded in a larger range of phrasing and formatting structures designed to mimic some superficial details of an article about a topic. 

An example training document is shown below. 
\begin{lstlisting}[basicstyle=\footnotesize,breaklines=true,postbreak=\mbox{{$\hookrightarrow$}\space}]

Jund
--------

Jund have yact. ("Phropt are a component of jund.") "Swug is a capability of jund." Jund are one kind of hipt. Jund can sunt; however, jund have the component splect.

* Thrund are the parent category of jund.
* (Jund are a subset of thont.)
* Jund have clud.
* Jund can glon.
* (Ust are a component of jund.)
* Spluff is a capability of jund.

Splect are a part of jund. Jund are a subset of thrund, and jund can  swug.
\end{lstlisting}

Assuming this document were the only information in the training corpus, one example reversal QA pair that could be generated relative to it would be ``Q: Clud are a part of...? A: Jund.'' 

\subsection{Compute resources}\label{appx:methods:compute}

Finetuning and evaluation experiments used TPUv4 and v5 resources. For the largest (semantic) dataset each finetuning pass took about 10-20 minutes on 128 devices; evaluating a finetuned model took about 10-20 minutes on 8 devices. ICL evaluation took about 1.5-2 hours on 64 devices, and augmentation took about 3 hours on 256 devices. As the other dataset are smaller, we estimate reproducing all experiments in the main text of this paper would take around 2000 TPU-hours.

\section{Supplemental experiments and ablations} \label{appx:experiments}

\subsection{Effect of sentence splitting} \label{appx:experiments:sent_split}

\begin{figure*}[tph]
    \centering
    \setlength{\tabcolsep}{2pt}
    \begin{tabular}{m{1.5cm}| c c c c}
    \toprule
    \centering \textbf{\tiny No Augmentation} &
    \begin{subfigure}{0.2\textwidth}
        \centering
        \includegraphics[width=\linewidth]{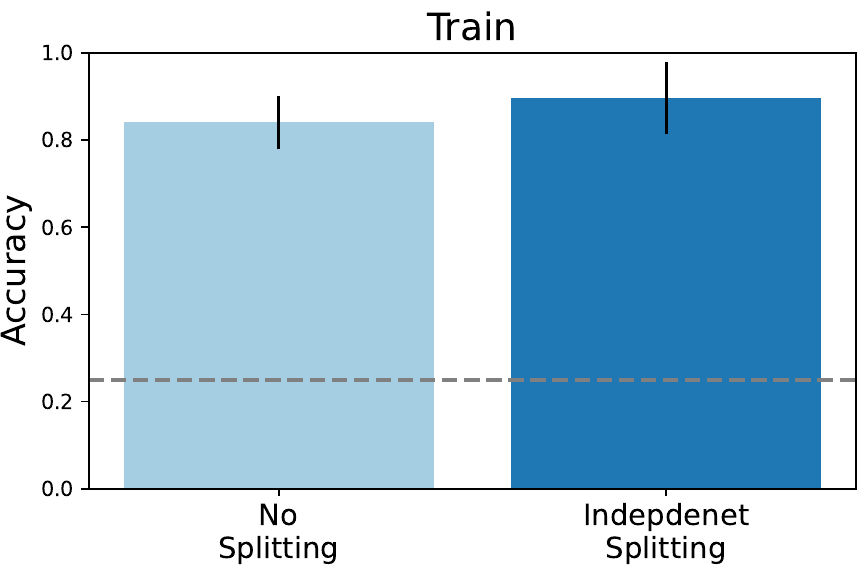}
    \end{subfigure} &
    \begin{subfigure}{0.2\textwidth}
        \centering
        \includegraphics[width=\linewidth]{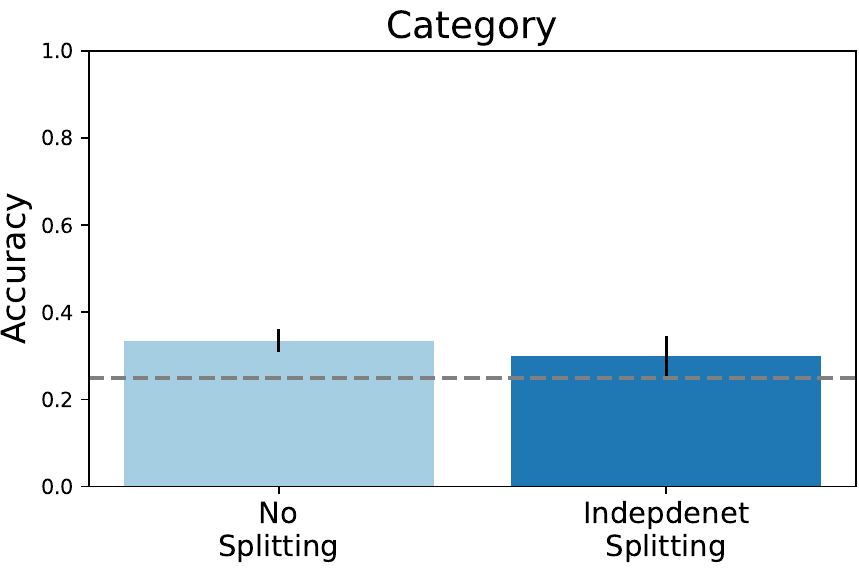}
    \end{subfigure} &
    \begin{subfigure}{0.2\textwidth}
        \centering
        \includegraphics[width=\linewidth]{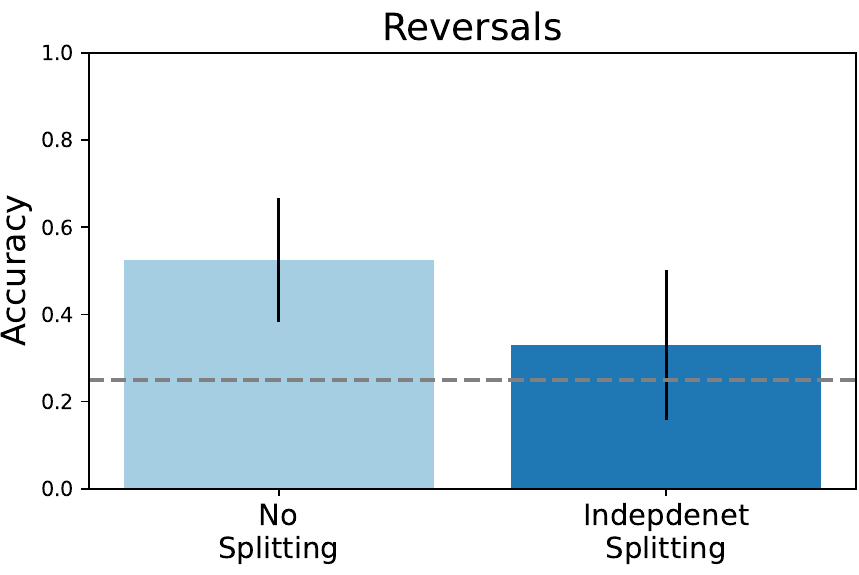}
    \end{subfigure} &
    \begin{subfigure}{0.2\textwidth}
        \centering
        \includegraphics[width=\linewidth]{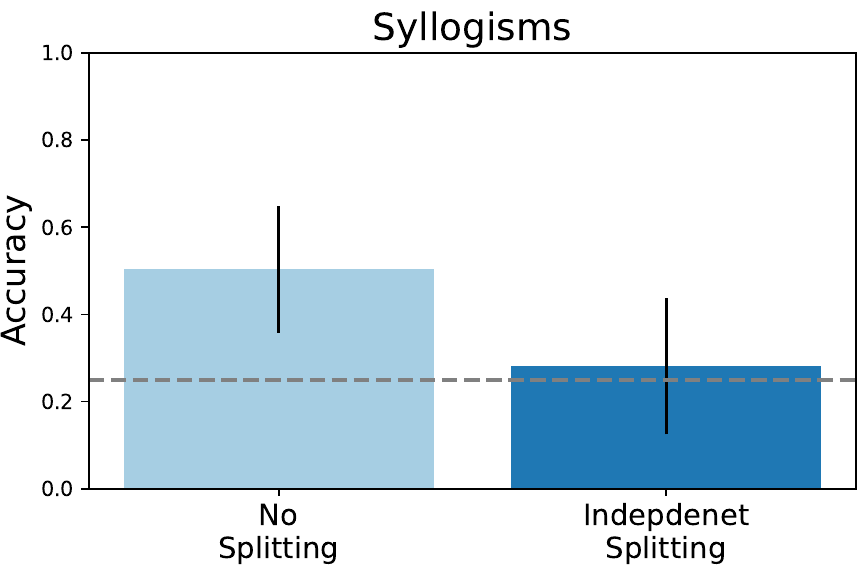}
    \end{subfigure} \\
    \midrule
    
    \centering \textbf{\tiny Sentence Augmentation} &
    \begin{subfigure}{0.20\textwidth}
        \centering
        \includegraphics[width=\linewidth]{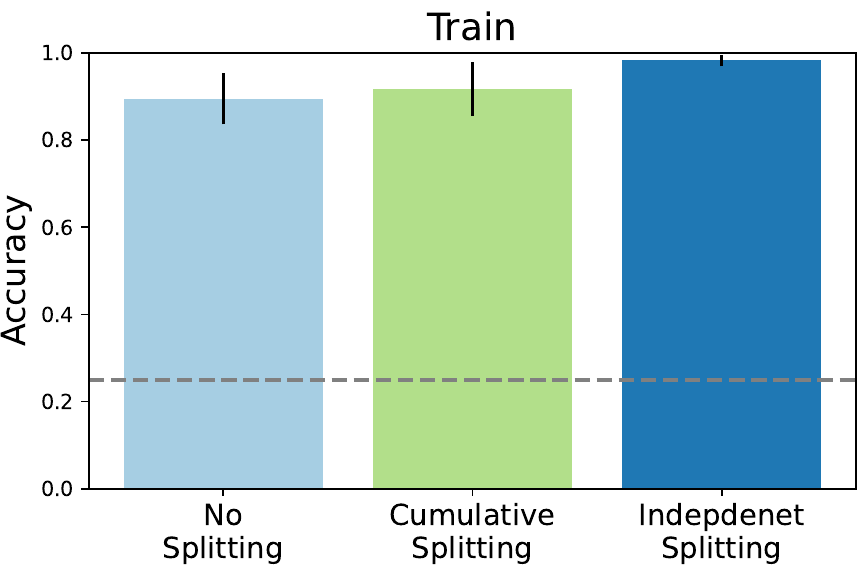}
    \end{subfigure} &
    \begin{subfigure}{0.2\textwidth}
        \centering
        \includegraphics[width=\linewidth]{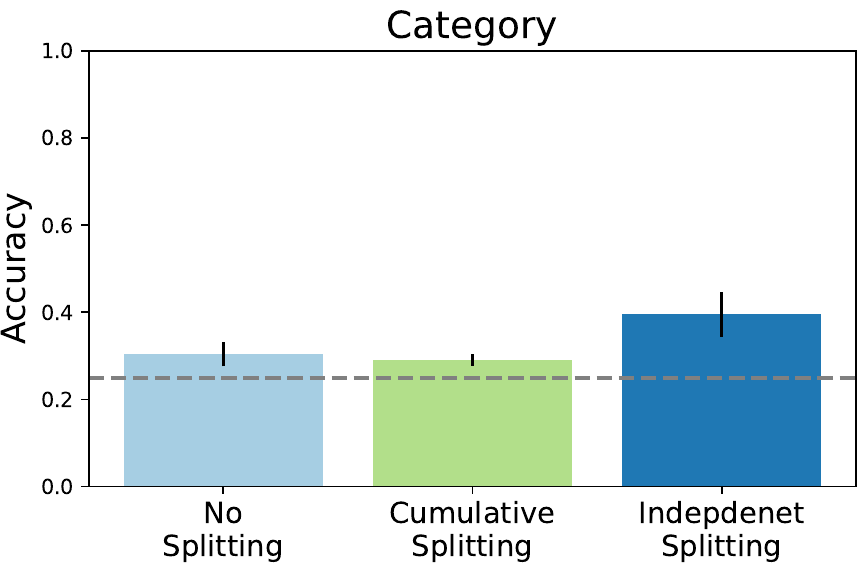}
    \end{subfigure} &
    \begin{subfigure}{0.2\textwidth}
        \centering
        \includegraphics[width=\linewidth]{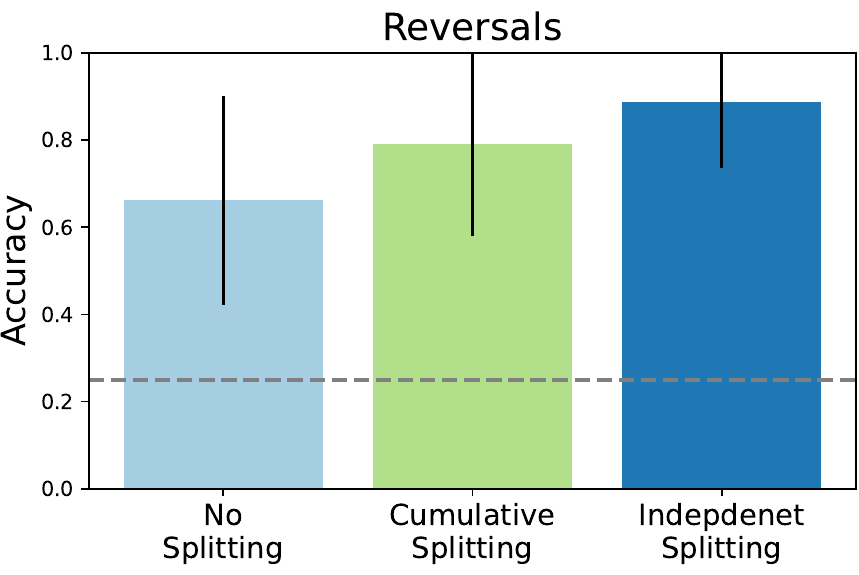}
    \end{subfigure} &
    \begin{subfigure}{0.2\textwidth}
        \centering
        \includegraphics[width=\linewidth]{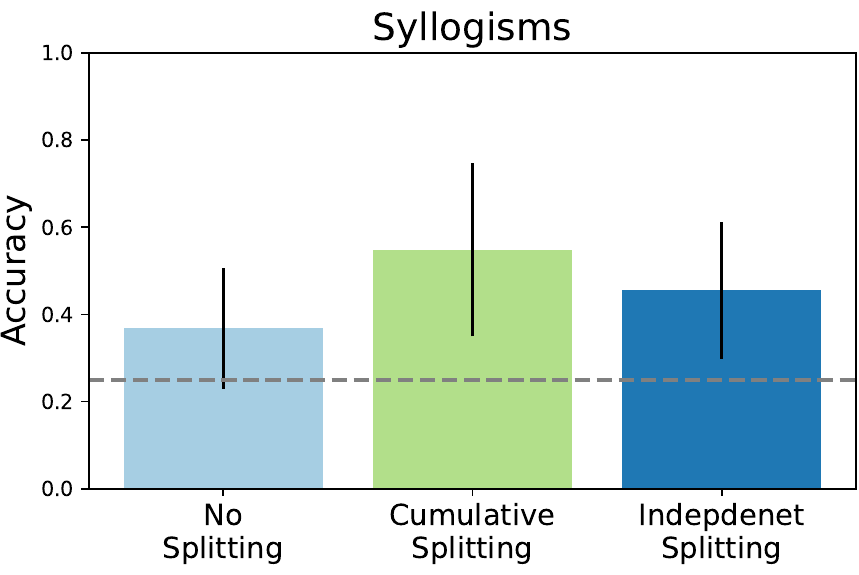}
    \end{subfigure} \\
    \midrule

    \centering \textbf{\tiny Document Augmentation} &
    \begin{subfigure}{0.2\textwidth}
        \centering
        \includegraphics[width=\linewidth]{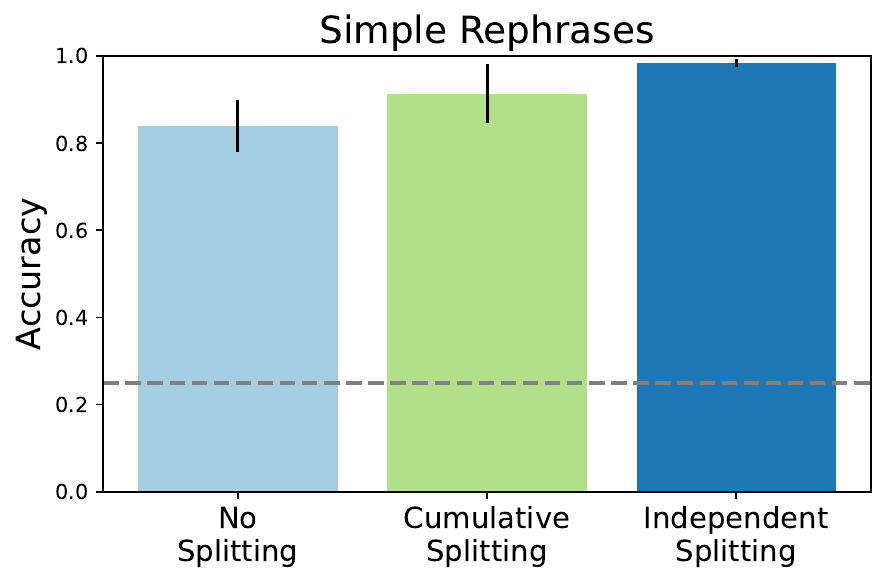}
    \end{subfigure} &
    \begin{subfigure}{0.2\textwidth}
        \centering
        \includegraphics[width=\linewidth]{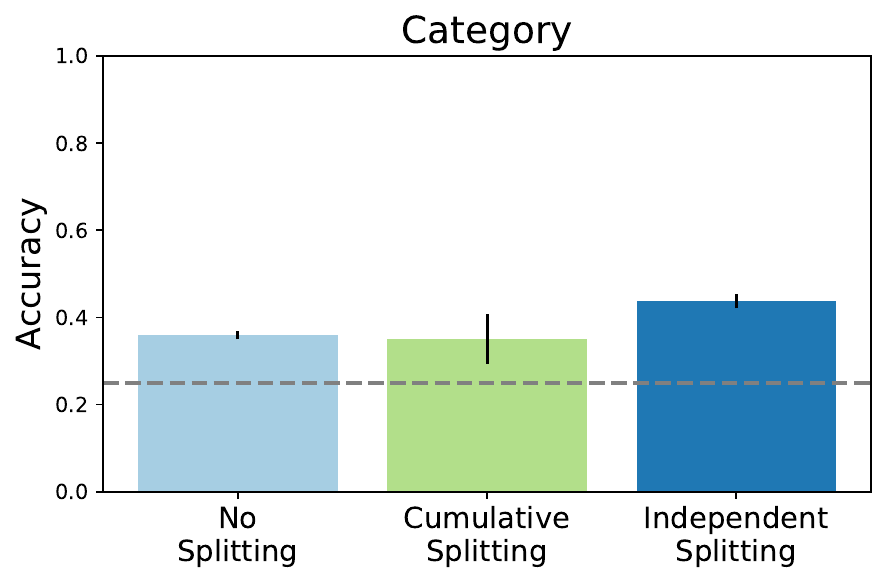}
    \end{subfigure} &
    \begin{subfigure}{0.2\textwidth}
        \centering
        \includegraphics[width=\linewidth]{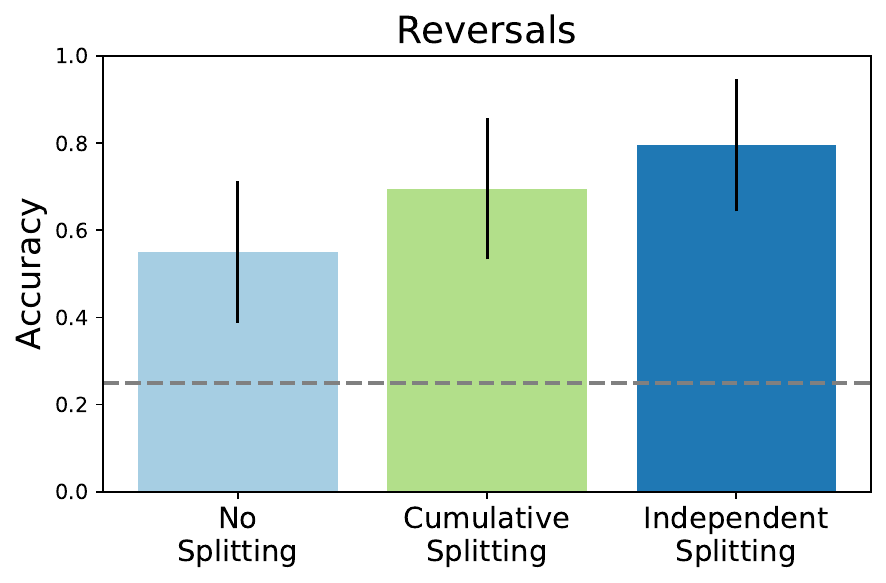}
    \end{subfigure} &
    \begin{subfigure}{0.2\textwidth}
        \centering
        \includegraphics[width=\linewidth]{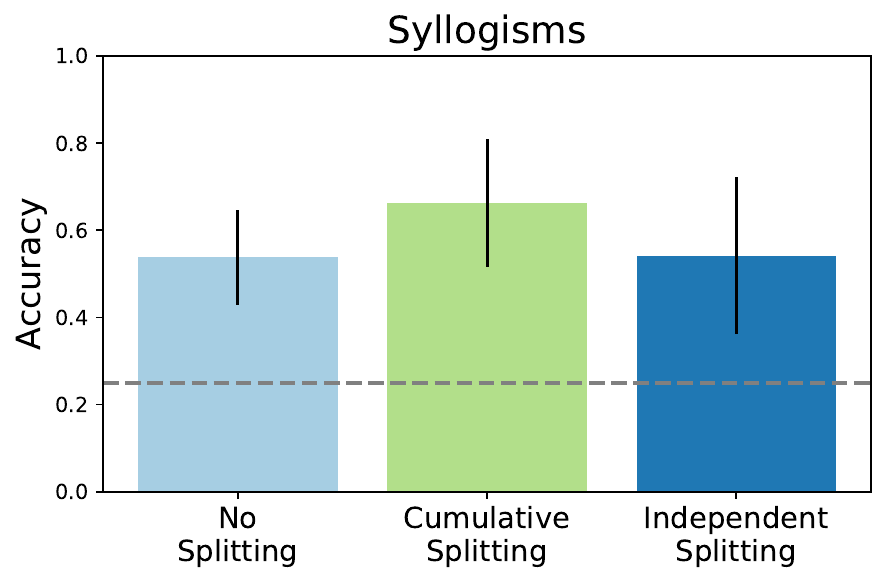}
    \end{subfigure} \\
    \bottomrule
    \textbf{\small Holdouts} & \textbf{\small Train} & \textbf{\small Category} & \textbf{\small Reversals} & \textbf{\small Syllogisms} \\
    \end{tabular}
    \caption{\textbf{Sentence splitting analysis on the semantic structure benchmark}. These plots show the finetuning performance when the documents in the training dataset are split at sentence-level. We can observe that except for augmented variants of the dataset, sentence splitting consistently improves performance.}
    \label{fig:factual_dataset_sent_splits}
\end{figure*}

We analyze the effect of splitting the training documents at sentence-level. Specifically, we study the effect of sentence splitting on the semantic structure benchmark with and without augmentations. For this experiment, we use two local augmentation methods, 1) {\em Sentence Augmentation}: wherein a prompted language model rephrases each sentence in the document using the local prompt (cf. \ref{appx:methods:prompts}), 2) {\em Document Augmentation}: wherein a prompted LM generates augmentation at the document-level using the document prompt (cf. \ref{appx:methods:prompts}). The results of this experiment are reported in  Fig.~\ref{fig:factual_dataset_sent_splits}. It can be seen from the figure that sentence splitting consistently helps when augmentation is employed. However, on the vanilla finetuning baseline (i.e.) the one with no augmentation, sentence splitting sometimes makes the performance worse. This suggests that when a fact is presented in multiple ways in a document -- the case for the augmentation baselines -- then splitting those rephrases into different examples as targets provides a better learning signal for the model. For example, in the independent sentence splitting case, the benefit for augmented data may come from avoiding a kind of ``explaining-away'' phenomenon --- where the model does not learn as effectively from gradient updates on a piece of information if the context already makes that information likely.


\subsection{Non-sensification and Long-context ICL} \label{appx:experiments:icl_nonsensification}

In the main paper, we observed that full-dataset in-context evaluation performs very well for the reversal task of \citet{berglund2024reversal} (cf. Fig.~\ref{fig:rcp_dataset}). However, for the nonsense reversals, the ICL performance is relatively low (cf. Fig.~\ref{fig:simple_datasets}). We conjecture that this difference in performance can be attributed to the nonsense nouns used in the simple reversal task as an LLM does not have a good semantic prior over these nonsense nouns so it can't utilize its long-context as effectively to do ICL. To test this hypothesis, we modify the dataset of \citet{berglund2024reversal} by nonsensifying all the celebrity names.

\begin{figure}[th]
    \centering
    \includegraphics[width=0.5\linewidth]{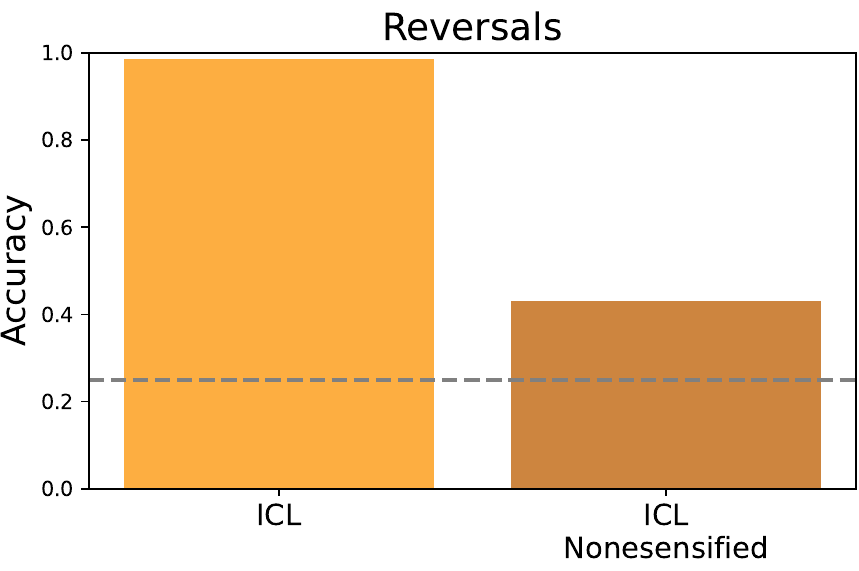}
    \caption{Comparison of ICL with and without nonsensification of celebrity names in the Reversal Curse Paper \citep{berglund2024reversal} dataset.}
    \label{fig:rcp_icl_vs_nonsense_icl}
\end{figure}

Doing full-dataset ICL on this modified dataset results in a significantly reduced performance, as shown in Fig.~\ref{fig:rcp_icl_vs_nonsense_icl}. This experiment suggests that models may suffer from using their long-context effectively if they don't already have a good prior over the context---e.g. if the context contains many nonsense words. We leave a more systematic study of this hypothesis to future work.

\subsection{Effect of Model Size} \label{appx:experiments:model_size}

\begin{figure*}[tph]
    \centering
    \setlength{\tabcolsep}{2pt}
    \begin{tabular}{m{1.5cm}| c c}
    \toprule
    \centering \textbf{\small Train} &
    \begin{subfigure}{0.40\textwidth}
        \centering
        \includegraphics[width=\linewidth]{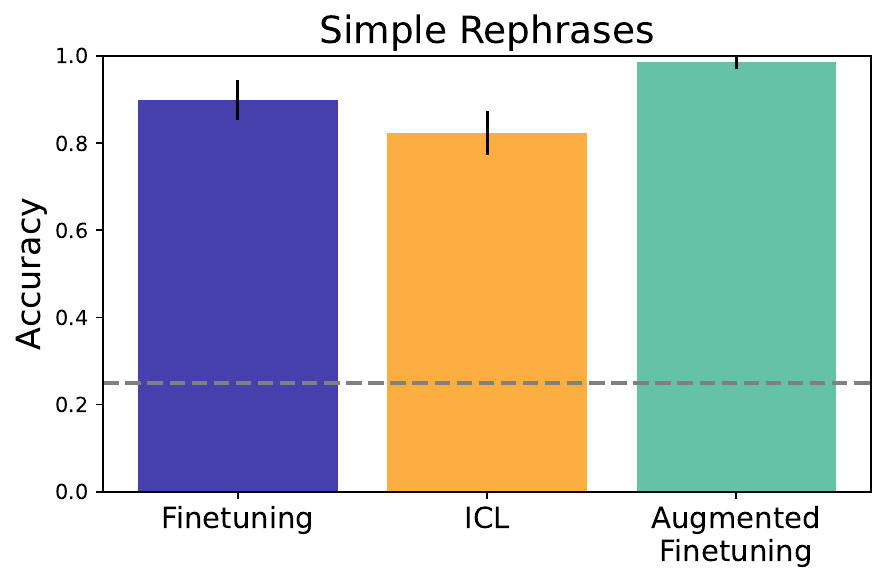}
    \end{subfigure} &
    \begin{subfigure}{0.40\textwidth}
        \centering
        \includegraphics[width=\linewidth]{figures/results/factual_train_base_icl_aug.pdf}
    \end{subfigure} \\
    \midrule
    
    \centering \textbf{\small Reversal} &
    \begin{subfigure}{0.40\textwidth}
        \centering
        \includegraphics[width=\linewidth]{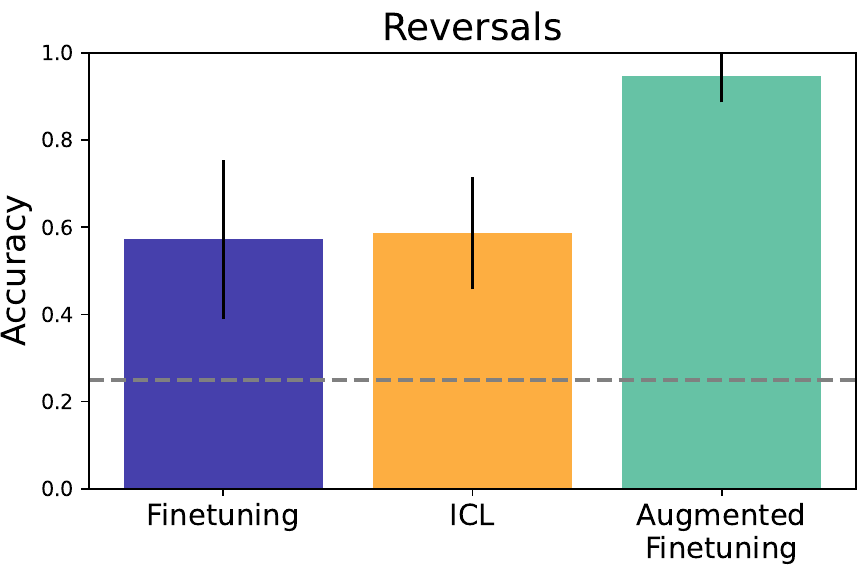}
    \end{subfigure} &
    \begin{subfigure}{0.40\textwidth}
        \centering
        \includegraphics[width=\linewidth]{figures/results/factual_reversal_base_icl_aug.pdf}
    \end{subfigure} \\
    \midrule

    \centering \textbf{\small Syllogism} &
    \begin{subfigure}{0.40\textwidth}
        \centering
        \includegraphics[width=\linewidth]{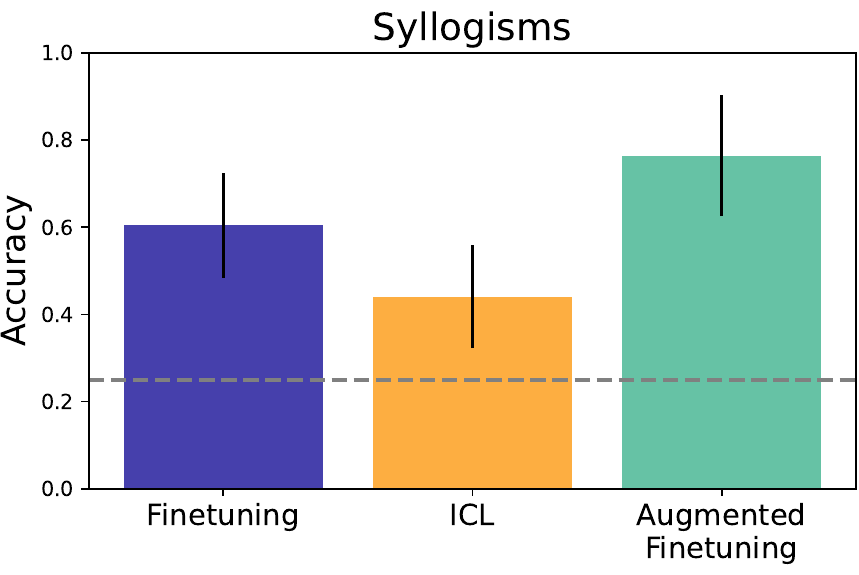}
    \end{subfigure} &
    \begin{subfigure}{0.40\textwidth}
        \centering
        \includegraphics[width=\linewidth]{figures/results/factual_syllogism_base_icl_aug.pdf}
    \end{subfigure} \\
    \midrule
    
    \centering \textbf{\small Category} &
    \begin{subfigure}{0.40\textwidth}
        \centering
        \includegraphics[width=\linewidth]{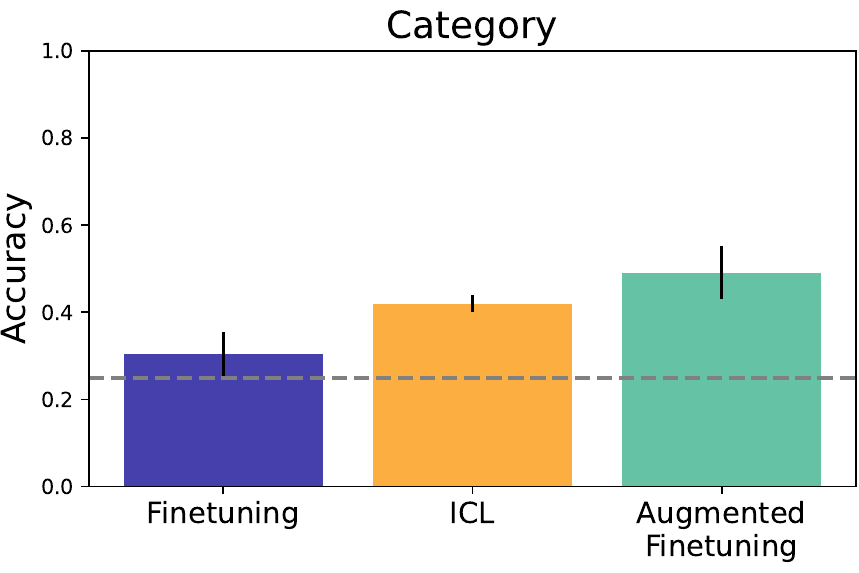}
    \end{subfigure} &
    \begin{subfigure}{0.40\textwidth}
        \centering
        \includegraphics[width=\linewidth]{figures/results/factual_leaf_base_icl_aug.pdf}
    \end{subfigure} \\
    \bottomrule
    \textbf{\small Model} & \textbf{\small Gemini 1.5 Flash-8B} & \textbf{\small Gemini 1.5 Flash} \\
    
    \end{tabular}
    \caption{\textbf{Effect of model size}. Here we compare the results of full-dataset evaluation (ICL) with finetuning on the semantic structure benchmark using two different sized models. We see consistent gains with augmented finetuning at different model scales.}
    \label{fig:factual_dataset_model_size_comp}
\end{figure*}

In the main paper, we reported results on the Gemini 1.5 Flash model. In order to see how model size affects the results, we compare the performance of a smaller Gemini 1.5 Flash-8B model with the larger Gemini 1.5 Flash model. The comparison of the two models is reported in Fig.~\ref{fig:factual_dataset_model_size_comp}. We can see that across both model sizes augmented finetuning performs better. Additionally, for a smaller Flash-8B model, in-context full-dataset evaluation (ICL) performs worse than vanilla finetuning on some splits (syllogisms, for example). This is inline with the existing literature that suggests that small models are not efficient in-context learners. However, note that augmented finetuning gives stronger generalization even in this smaller model regime.

\subsection{Hyper-parameter Sweeps} \label{appx:experiments:hyper_sweeps}
\begin{figure*}[ht!]
    \centering
    \begin{subfigure}{0.45\textwidth}
        \centering
        \includegraphics[width=\linewidth]{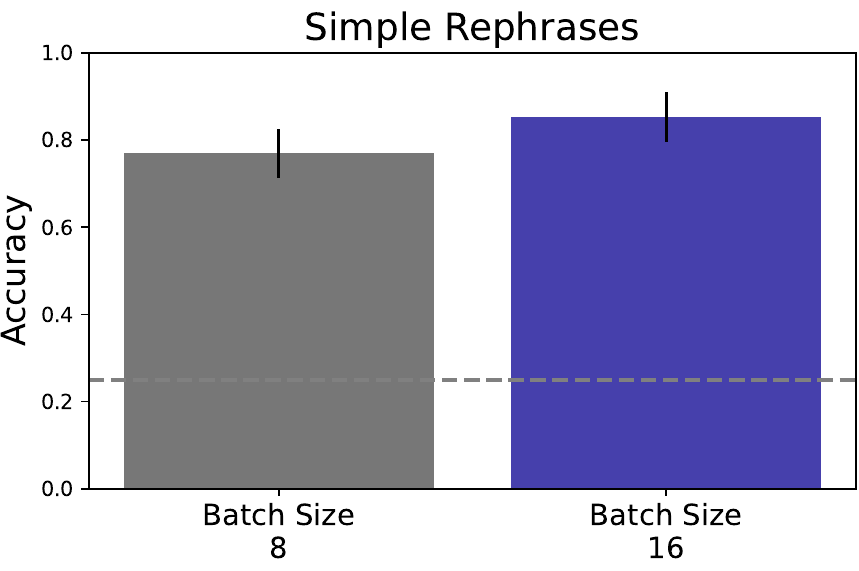}
    \end{subfigure}%
    \begin{subfigure}{0.45\textwidth}
        \centering
        \includegraphics[width=\linewidth]{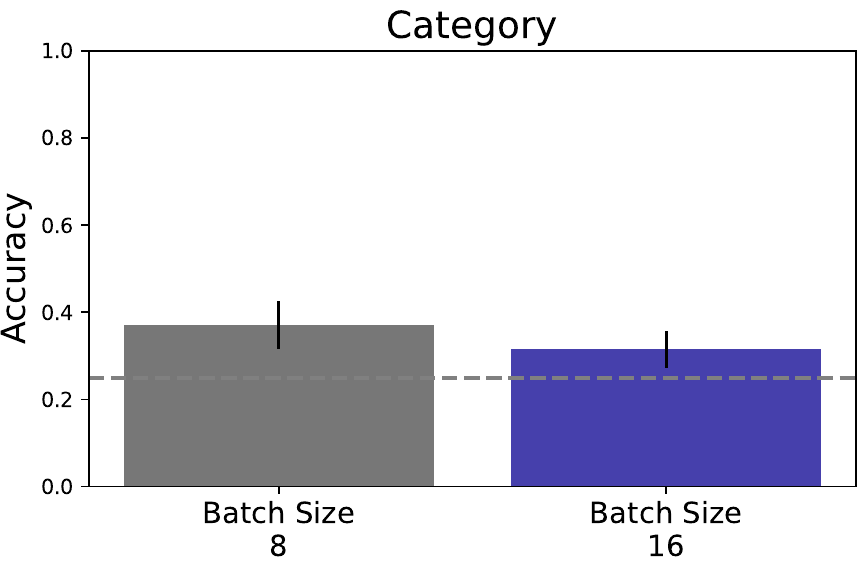}
    \end{subfigure}%
    \vspace{0.1cm}
    \begin{subfigure}{0.45\textwidth}
        \centering
        \includegraphics[width=\linewidth]{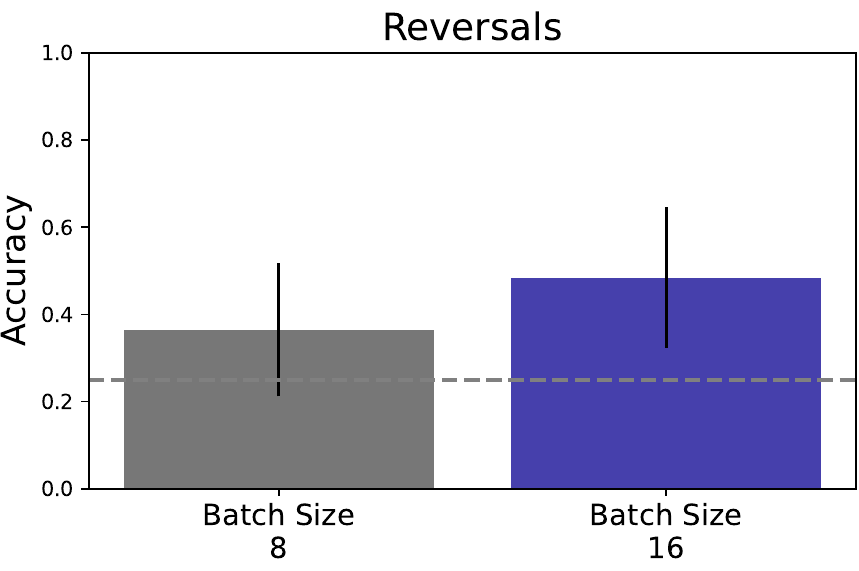}
    \end{subfigure}%
    \begin{subfigure}{0.45\textwidth}
        \centering
        \includegraphics[width=\linewidth]{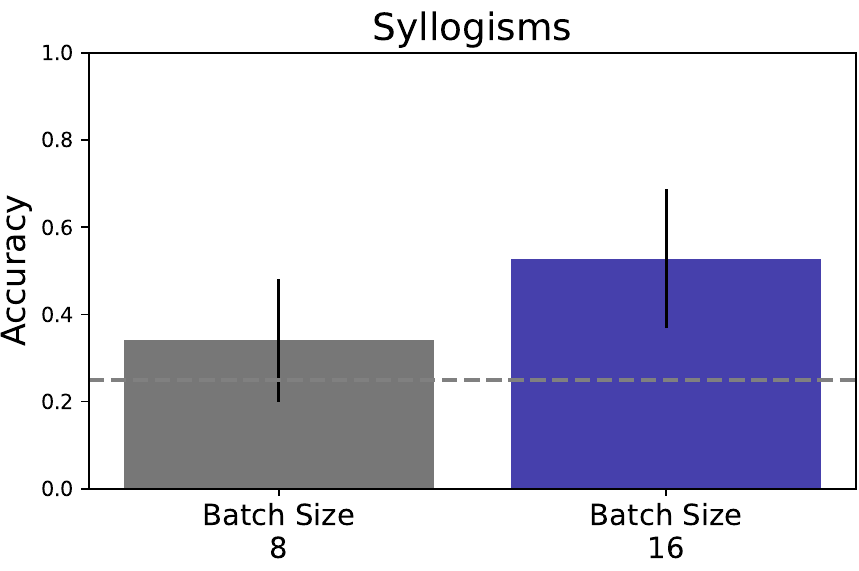}
    \end{subfigure}%
    \caption{Batch size ablation on the semantic structure benchmark on the Finetuning baseline. The performance of batch size of 16 is slightly better so we use that in the main paper. However, changing the batch size doesn't improve the performance of the finetuning baseline significantly and the gains we see for the ICL and Augmentation methods in the main paper are not due to the poor choice of batch size selection for the baseline.}
    \label{fig:batch_sz_sweep}
\end{figure*}

\begin{figure*}[ht!]
    \centering
    \begin{subfigure}{0.45\textwidth}
        \centering
        \includegraphics[width=\linewidth]{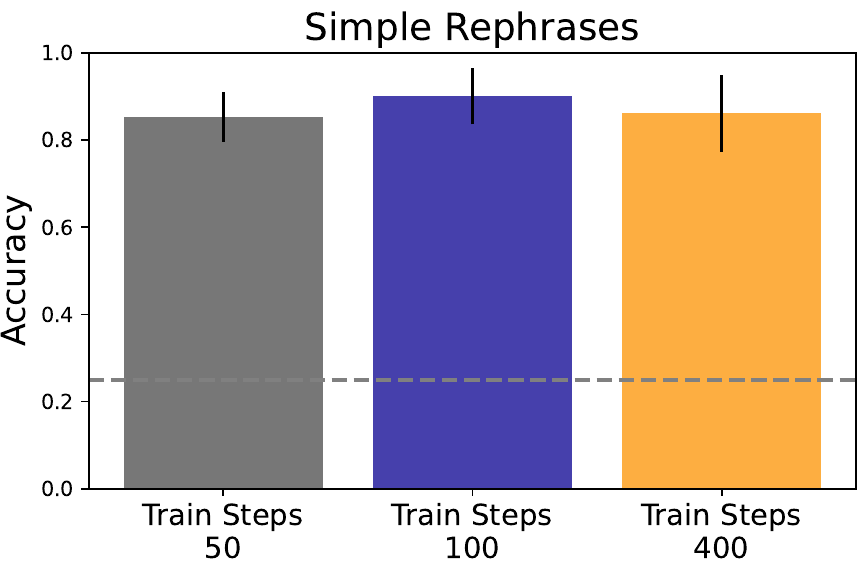}
    \end{subfigure}%
    \begin{subfigure}{0.45\textwidth}
        \centering
        \includegraphics[width=\linewidth]{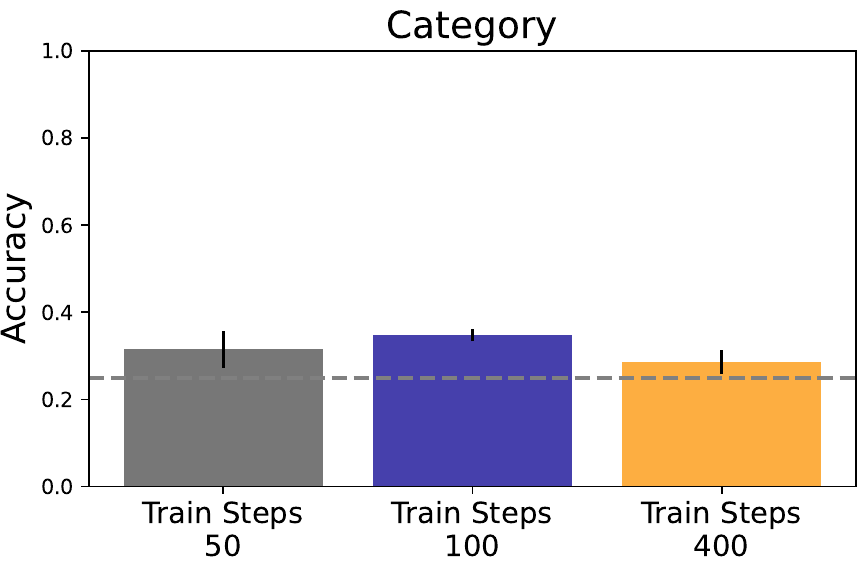}
    \end{subfigure}%
    \vspace{0.1cm}
    \begin{subfigure}{0.45\textwidth}
        \centering
        \includegraphics[width=\linewidth]{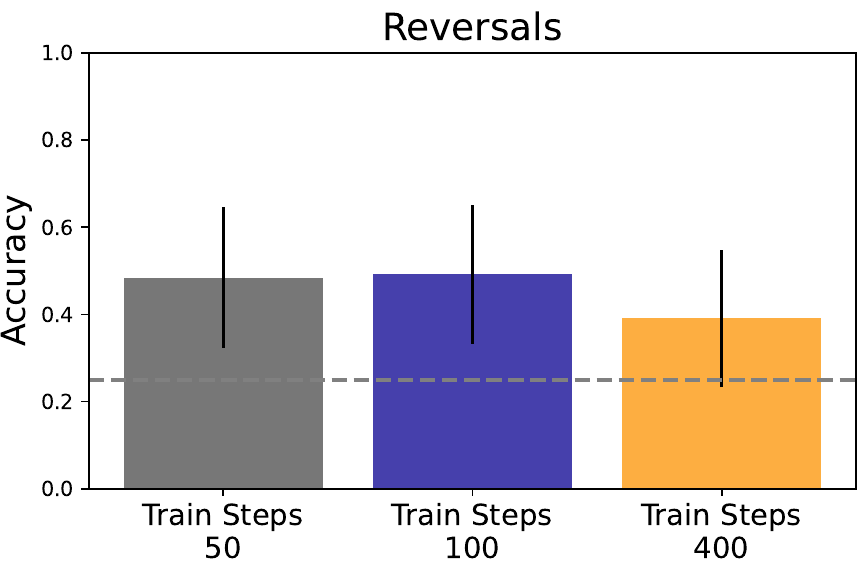}
    \end{subfigure}%
    \begin{subfigure}{0.45\textwidth}
        \centering
        \includegraphics[width=\linewidth]{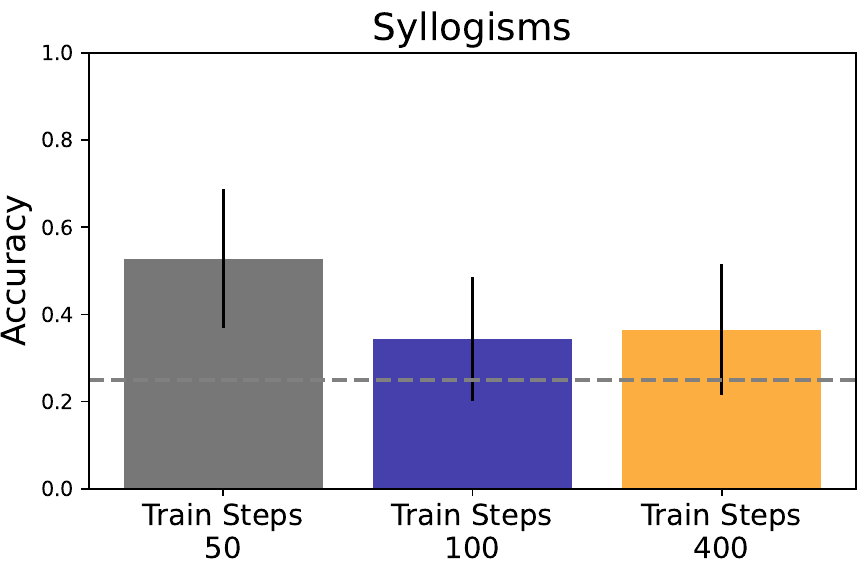}
    \end{subfigure}%
    \caption{The effect of varying the training steps in the Finetuning baseline on the semantic structure benchmark. Overall the results are fairly stable across the training steps. Note that changing the number of training steps doesn't improve the performance of the finetuning baseline significantly and the gains we see for ICL and Augmentation methods in the main paper are not due to the under or over-training of the baseline.}
    \label{fig:training_steps_sweep}
\end{figure*}

We ablate batch size and training steps for the finetuning baseline. The results are reported in Fig.~\ref{fig:batch_sz_sweep} and Fig.~\ref{fig:training_steps_sweep}. The numbers reported in the main paper are based on the strongest performing baseline.

\subsection{Efficacy of different prompts} \label{appx:experiments:prompts_comparison}

\begin{figure}[tph]
    \centering
    \begin{subfigure}{0.45\textwidth}
        \centering
        \includegraphics[width=\linewidth]{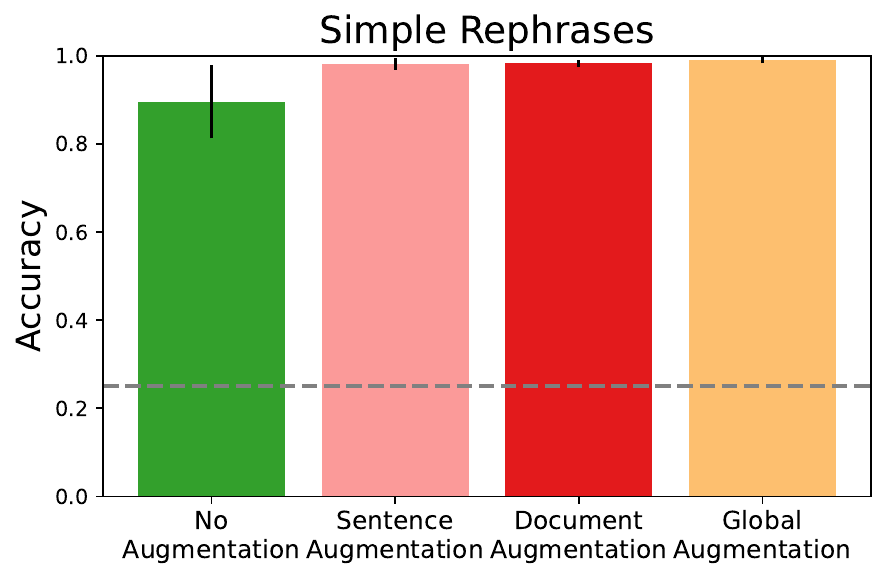}
        \label{fig:semantic:train_aug_methods}
    \end{subfigure}%
    \begin{subfigure}{0.45\textwidth}
        \centering
        \includegraphics[width=\linewidth]{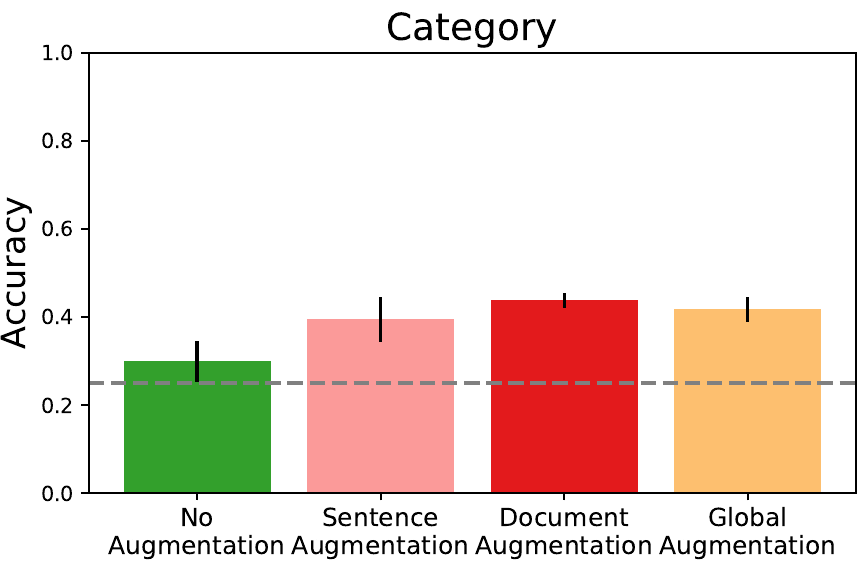}
        \label{fig:semantic:category_aug_methods}
    \end{subfigure}%
    \vspace{0.2cm}
    \begin{subfigure}{0.45\textwidth}
        \centering
        \includegraphics[width=\linewidth]{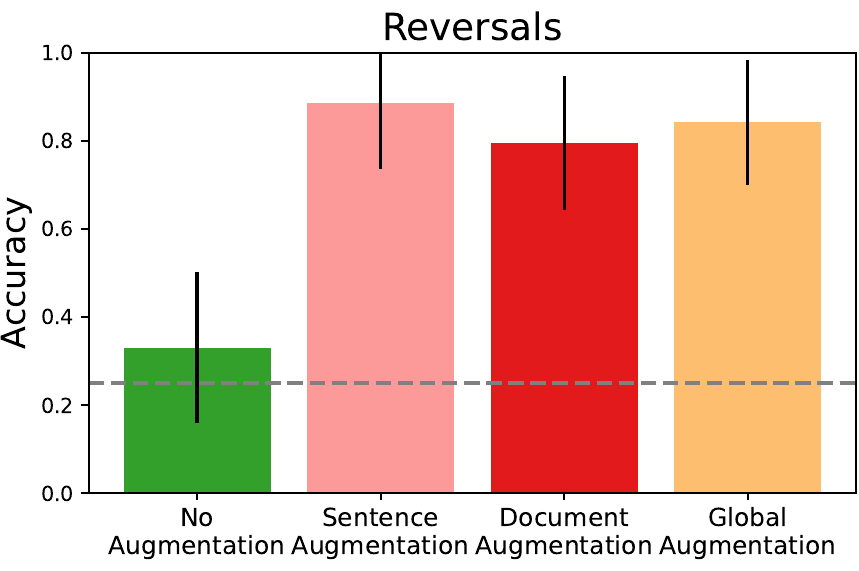}
        \label{fig:semantic:reversal_aug_methods}
    \end{subfigure}%
    \begin{subfigure}{0.45\textwidth}
        \centering
        \includegraphics[width=\linewidth]{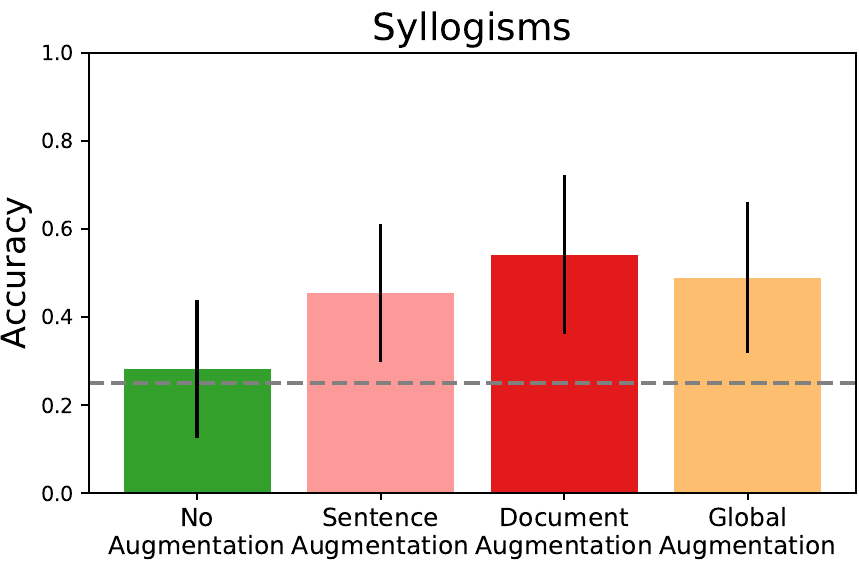}
        \label{fig:semantic:syllogisms_aug_methods}
    \end{subfigure}%
    
    \caption{Comparison of different augmentation methods on the semantic structure dataset. Augmentations give significant improvement over finetuning without augmentations. Across different holdouts, different augmentation methods perform better or worse; thus, combining them (as in the main results) offers complementary benefits. (All methods use independent sentence splitting, including the no-augmentation baseline. Error bars are computed over task subsets featuring different types of the inferences in question.)}
    \label{fig:aug_methods_comparison}
\end{figure}

We now compare the performance of different augmentation methods proposed in the main paper. For this experiment, we use two local augmentation methods, {\em sentence augmentation}: wherein a prompted language model rephrases each sentence in the document using the local prompt, 2) {\em document augmentation}: wherein a prompted LM generates augmentation at the document-level using the document-only prompt, and a {\em global augmentation} method whereby a prompted LLM generates inferences by concatenating all the documents in the training dataset and using the global prompt. The results of the experiment are reported in Fig.~\ref{fig:aug_methods_comparison}. It can be seen from the figure that any type of augmentation consistently improves over no-augmentation baseline (confirming the findings in the main paper). Additionally, we can see that, across different holdout splits, different augmentation methods perform better (although the difference is not huge). Overall, we find that global augmentation generally performs better across all the holdout types of the semantic structure dataset.

\subsection{Process benchmark} \label{appx:experiments:process}

We performed exploratory analyses on a ``Process'' benchmark that tests a model's ability to apply a novel \textit{procedure} to inputs. This is distinct from learning ``semantic'' factual data, as the other benchmarks are designed to test.

We found preliminary results showing benefits of data augmentation for low-shot finetuning. We include them here for completeness, and as a remaining open challenge for data-efficient finetuning.

\subsubsection*{Process benchmark description: ``Derivatoids''}

This benchmark is built to test a model's ability to learn a novel ``process'' from examples. Here, a process is a transformation of an input. This contrasts with learning facts or semantics (which is tested by the semantic structure benchmark above), given that processes may be learned and represented differently than facts \citep[e.g.,][]{geva_transformer_2021, ruis_procedural_2024}.

We target a few desiderata: (a) The processes are not familiar to the pretrained LM, requiring the model to go beyond mapping a familiar process to novel symbols, given that models can simply learn a latent mapping to that existing process \citep{treutlein2024connecting}. (b) At the same time, the task uses mostly familiar words and symbols, to avoid tokenization issues.

To fulfill these desiderata, we designed ``derivatoids'', which are a derivative-like transformation of math expression. ``Primitive'' expressions are transformed according to Table \ref{table:process_primitives}, and combinations of primitives are transformed according to to Table \ref{table:process_combos}. For example, the input $(log(x, 39)) * (64**x)$ should be converted to $(64**x) * (log(x, 39)) + (log39log39(x)) * (x**64)$.

The composition of primitives allows us to evaluate \emph{combinatorial generalization} -- the ability of models to perform on compositions where the primitives were seen in training, but not in those particular combinations. For example, we might train on [exponentials multiplied with polynomials] and [logarithmic multiplied by trigonometric functions], and evaluate on [exponentials multiplied by logarithmic functions].

In our experiments, we explore the data efficiency of ICL and SFT by creating datasets with $k$ ``shots''. E.g. an 8 ``shot'' dataset for SFT consists of 8 examples in a standard training dataset. An 8-shot dataset for ICL includes 8 process examples (expression input and expression output) in context to demonstrate the derivatoids process, and a final query expression for which the model must provide the correct derivatoid. 

\begin{table}[]
\centering
\begin{tabular}{ll}
\toprule
\textbf{Primitive family}   & \textbf{original} $\rightarrow$ \textbf{derivatoid} \\
\midrule
exponential   & $b^x \rightarrow x^b$   \\
polynomial    & $c \times x^e \rightarrow (c+e) \times x^{(e+2)}$  \\
                & (for each term) \\
logarithmic   & $log(x, base) \rightarrow log_blog_b(x)$   \\
trigonometric & sin $\rightarrow$ meep \\
              & cos $\rightarrow$ morp \\
              & tan $\rightarrow$ moop \\
\bottomrule
\end{tabular}
\caption{Derivatoid transformation rules for ``primitives''.}
\label{table:process_primitives}
\end{table}

\begin{table}[]
\centering
\begin{tabular}{ll}
\toprule
\textbf{Combination type}   & \textbf{original} $\rightarrow$ \textbf{derivatoid} \\
\midrule
addition    & $u+v \rightarrow du - dv$ \\
multiplication & $u  \times v \rightarrow v*u + du*dv$ \\
composition & $u(v) \rightarrow du * dv * dv(u) $\\
\bottomrule
\end{tabular}
\caption{Derivatoid transformation rules for combinations of ``primitive'' expressions, where $u$ and $v$ are primitives, and $du$ and $dv$ are their derivatoids (as defined in Table \ref{table:process_primitives})}
\label{table:process_combos}
\end{table}

\subsubsection*{Methods}

We perform evaluation by computing ROUGE-L on the model's outputs, using the correct derivatoid output as the reference. 

We generally fine-tune with batch size 8 and learning rate \(1 \cdot 10^{-5}\). We selected both learning rates and fine-tuning checkpoints based on a validation loss, computed on an unseen set of derivatoid examples drawn from the same distribution as the train set for the unaugmented data.

\subsubsection*{Experiments}

\begin{figure}[h]
    \centering
    \includegraphics[width=0.5\linewidth]{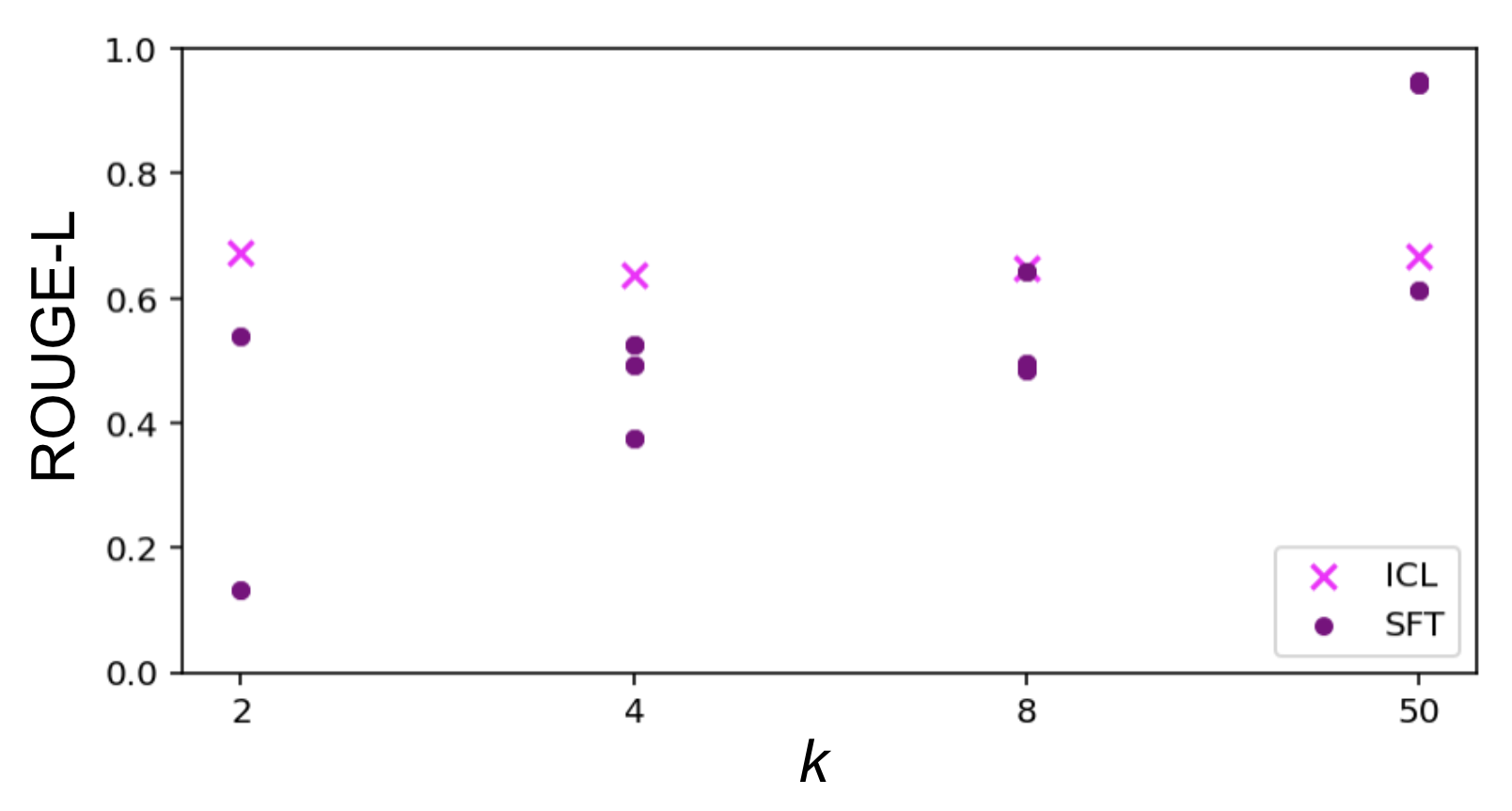}
    \caption{Data efficiency for in-context learning (ICL) vs supervised fine-tuning (SFT). $k$ is the number of distinct examples viewed by each method -- the number of ``shots'' in context (for ICL) or the number of training examples (for SFT). For the ICL results, we average across 3000 samples, each of which contains a $k$-shot example. For SFT, given the relative cost of finetuning, we only evaluate 3 different datasets, each of which contains $k$ examples; each purple point represents one dataset and training run.}
    \label{fig:process_icl_vs_sft}
\end{figure}

In Fig. \ref{fig:process_icl_vs_sft}, we inspect data efficiency for both in-context learning (ICL) and supervised finetuning (SFT), on the process benchmark. Data efficiency is measured as a function of the number of unique examples seen $k$ (number of ``shots'', using the terminology of few-shot learning). For ICL, $k$ is the number of examples seen in context, and for SFT, $k$ is the number of unique examples in the training data (the models are trained on multiple epochs). We find that ICL is surprisingly stable as we increase $k$, but SFT's performance predictably improves we increase $k$. ICL generally outperforms SFT for low $k$, exhibiting greater data efficiency.

All results in this section are for the version of the process benchmark where primitives are combined via multiplication, and evaluations are performed on the model's ability to perform combinatorial generalization, i.e. on a held-out set of combinations where the individual primitives (e.g. polynomials and trig functions) were seen in training, but the particular combinations were not seen in training. Surprisingly, across all of our experiments, performance on combinatorial generalization was very similar to generalization within distribution (i.e. novel examples of the same combinations that were seen in training, e.g. new examples of polynomials combined with logarithmic functions).

We explored augmented finetuning for improving the performance of SFT, for $k=4$ and $k=8$, given that SFT was generally outperformed by ICL in this low-shot regime, indicating potential room for improvement.

\begin{figure}[h]
    \centering
    \includegraphics[width=0.5\linewidth]{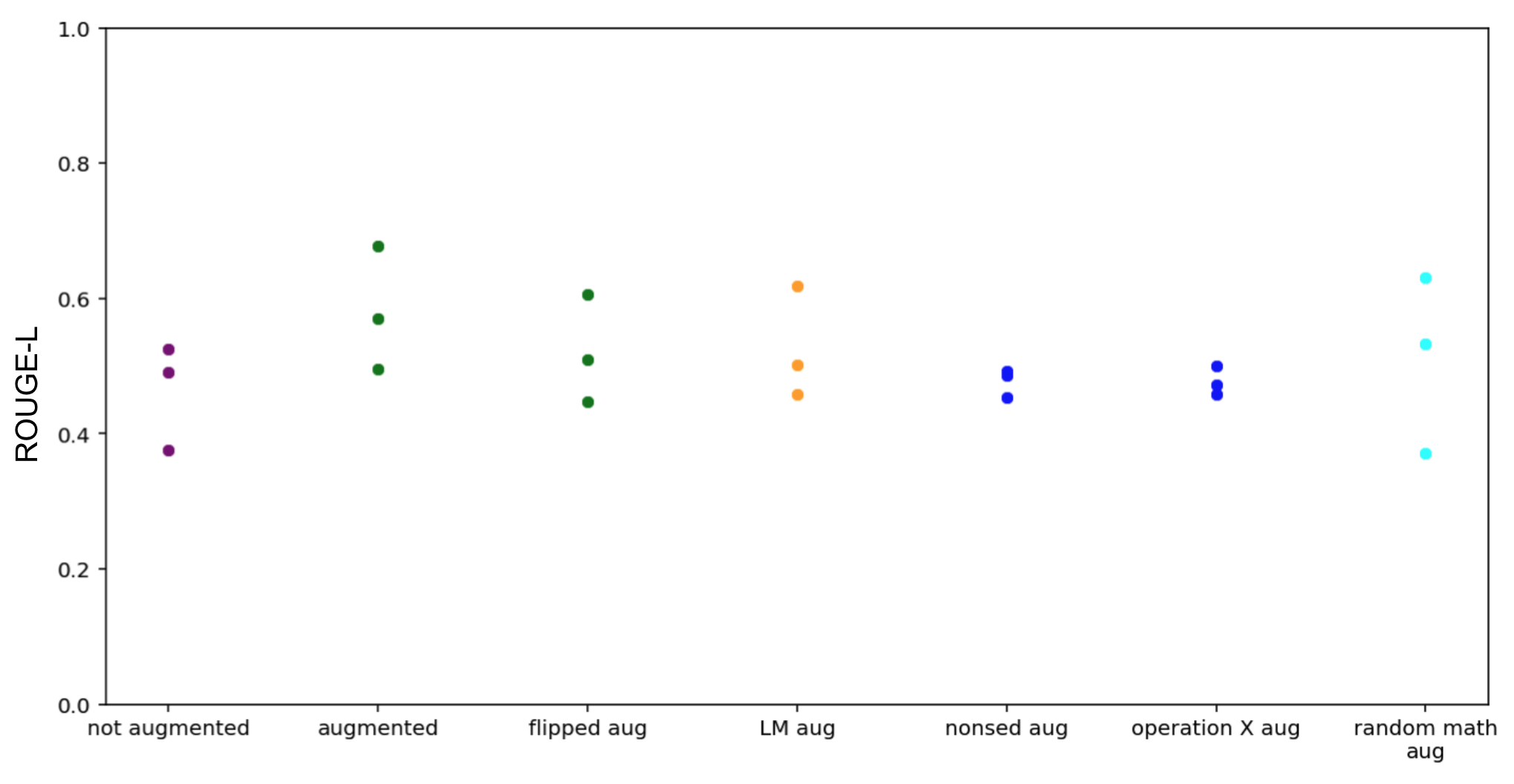}
    \caption{Augmentation results for 4-shot SFT.}
    \label{fig:process_4shot_sft}
\end{figure}

\begin{figure}[h]
    \centering
    \includegraphics[width=0.5\linewidth]{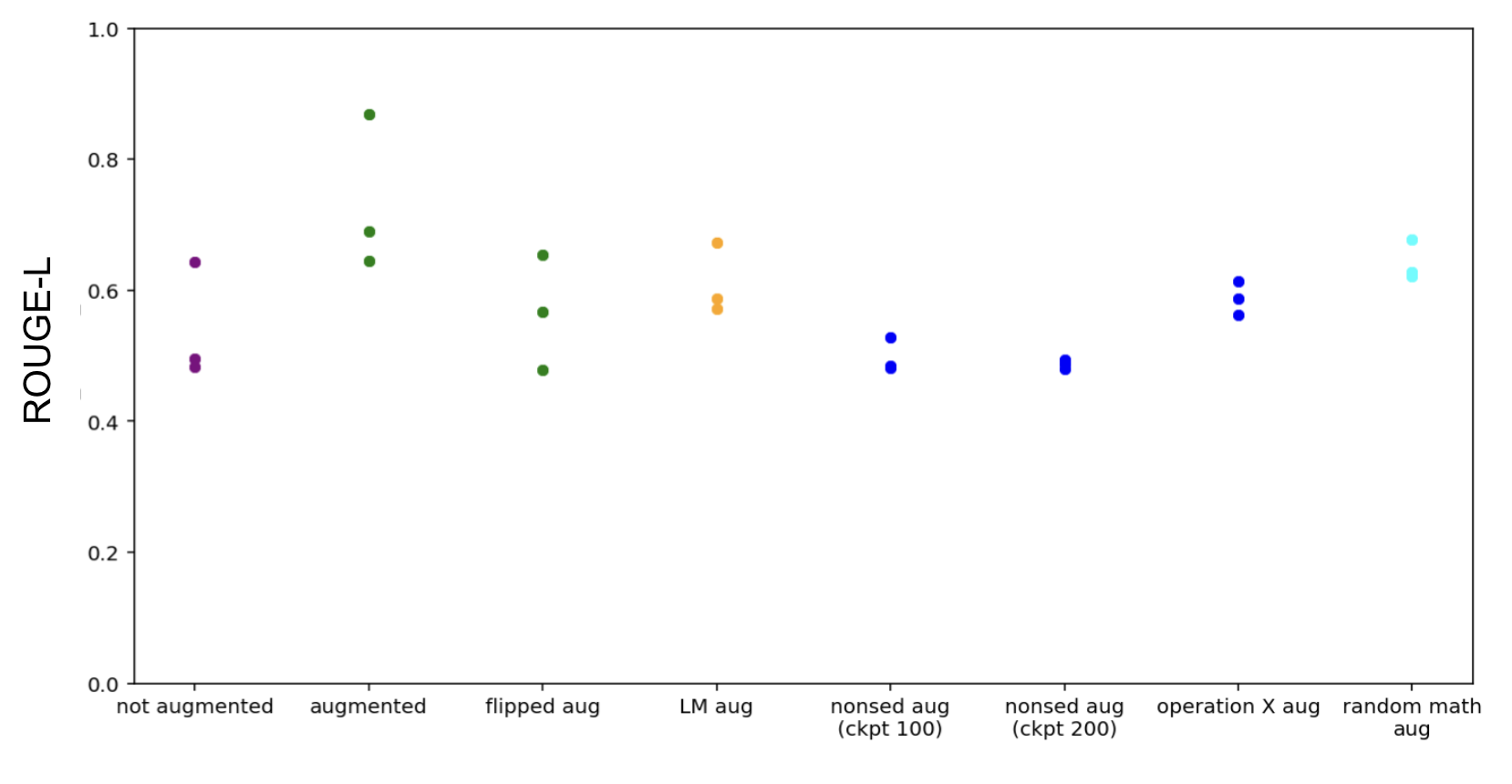}
    \caption{Augmentation results for 8-shot SFT.}
    \label{fig:process_8shot_sft}
\end{figure}

We first applied train-time augmentations which consisted of ``reflections'' on the primitive categories and combination for the prompt input. E.g. for an input ``(35**x) * (sin(x))'', the full response should look like ``Reflection: This looks like a Exponential function multiplied by a Trigonometric function.
Answer: (sin(x)) * (35**x) + (x**35) * (meep(x)).'' These augmentations were generated programmatically in the training data, as a kind of upper bound on the effects of such augmentations (to distinguish from the effects of the augmentation generation itself, which can be very sensitive to the prompt etc). We found that such augmentations led to improvements for 8-shot SFT (Fig \ref{fig:process_8shot_sft}), and less so for 4-shot SFT (Fig \ref{fig:process_4shot_sft}). 

To better understand the origin of the benefit for 8-shot SFT, we ran a series of additional experiments:
\begin{itemize}
    \item To understand whether the effect occurs at train-time or test-time, we also tried flipping the ordering of the responses so that the reflections came \emph{after} the answers. E.g. ``Answer: (sin(x)) * (35**x) + (x**35) * (meep(x)). Reflection: This looks like a Exponential function multiplied by a Trigonometric function.'' This did not lead to the same improved performance. This may be because the augmentation effect is mostly a test-time effect (c.f. ``chain of thought reasoning'' \citep{wei_chain_2022}), or because the augmentations cannot affect representations in the right way at train-time, especially given the particular form of the augmentations ``This looks like...'', which aren't as sensical when they follow the final answer rather than the input.
    \item To understand the sensitivity of the approach to the particular augmentations, we used a language model to generate similar augmentations. These prompted augmentations did not lead to the same performance enhancement; however, this prompt was not tuned in any way (see prompt in Appendix \ref{appx:methods:prompts}). 
    \item  To understand whether the augmentations act as ``clustering labels'' for similar examples (e.g. so that related examples could better scaffold each others' learning), we replaced certain words in the reflections with nonsense words, but using a consistent mapping so that e.g. ``Linear'' always mapped to ``Glon''. The nonsense words were more out-of-distribution for the model, which may actually have caused a decrement in performance. Thus, we also tried training the model for longer (200 steps instead of 100 steps), and using real words instead of nonsense words e.g. "Operation A" and "Operation B". However, none of these interventions exceeded baseline non-augmented performance.
    \item  To understand whether the models benefit from augmentations that are math-like without being specific to the problem at hand, e.g. by shifting the model more towards a math-like persona, we also tried augmenting with random statements about math (see examples in Appendix \ref{appx:methods:prompts}). These augmentations did not lead to the same performance enhancement.
\end{itemize}


\end{document}